\documentclass{applemlr}
\usepackage{amsmath}
\usepackage{enumerate}
\usepackage{algorithm}
\usepackage{algpseudocode}
\usepackage{amsfonts}
\usepackage{amsthm}
\usepackage{cleveref}
\usepackage{diagbox}
\usepackage{colortbl}
\usepackage{amssymb}
\usepackage{xspace}
\usepackage{wrapfig}
\usepackage{adjustbox}
\usepackage{tabularx}
\usepackage{booktabs}
\usepackage{mathtools}
\usepackage{tikz}
\usepackage{enumitem}
\usepackage{silence}
\usepackage{dsfont}
\usepackage[table]{xcolor}
\usepackage[dvipsnames]{xcolor}
\usepackage{multirow}
\usepackage{makecell}
\usepackage{xfakebold}

\usepackage{amsmath,amsfonts,bm}

\def\eqref#1{equation~\ref{#1}}

\def\1{\bm{1}}

\DeclareMathAlphabet{\mathsfit}{\encodingdefault}{\sfdefault}{m}{sl}
\SetMathAlphabet{\mathsfit}{bold}{\encodingdefault}{\sfdefault}{bx}{n}

\definecolor{textgray}{HTML}{6E6E73}
\usetikzlibrary{positioning, calc}
\usetikzlibrary{decorations.pathmorphing}

\makeatletter
\patchcmd{\wrong@fontshape}{\@gobbletwo}{}{}{}
\makeatother
\numberwithin{equation}{section}
\makeatletter
\AtBeginDocument{
  \urlstyle{sf}
  
}
\makeatother

\definecolor{light}{RGB}{125, 125, 125}
\crefname{tcb@cnt@pbox}{code}{code}
\Crefname{tcb@cnt@pbox}{Code}{Code}
\crefname{assumption}{assumption}{assumption}
\Crefname{assumption}{Assumption}{Assumptions}

\newtcolorbox[auto counter]{pbox}[2][]{
  colback=white,
  title=Code~\thetcbcounter: #2,
  #1,fonttitle=\sffamily,
  fontupper=\sffamily,
  arc=2pt,
  colframe=bgcolor,
  coltitle=fgcolor,
  colbacktitle=bgcolor,
  toptitle=0.25cm,
  bottomtitle=0.125cm
}

\makeatletter
\newcommand\applefootnote[1]{%
  \begingroup
  \renewcommand\thefootnote{}%
  \renewcommand\@makefntext[1]{\noindent##1}%
  \footnote{#1}%
  \addtocounter{footnote}{-1}%
  \endgroup
}
\makeatother

\definecolor{cverbbg}{gray}{0.90}

\usepackage{amsthm}

\usetikzlibrary{arrows.meta, positioning, calc, decorations.pathreplacing,
                shapes.geometric, fit, backgrounds, patterns, decorations.markings}
\usepackage{pgfplots}
\pgfplotsset{compat=1.17}

\usepackage{subcaption}
\usepackage{wrapfig}
\usepackage{multirow}
\usepackage{colortbl}
\usepackage{pifont}
\usepackage{float}
\usepackage{nicefrac}

\usepackage{xspace}
\usepackage{etoc}

\crefname{prop}{proposition}{propositions}
\Crefname{prop}{Proposition}{Propositions}
\AtBeginDocument{}

\providecommand{\REQUIRE}{\Require}

\providecommand{\STATE}{\State}

\providecommand{\FOR}[1]{\For{#1}}

\providecommand{\ENDFOR}{\EndFor}

\definecolor{dpsblue}{HTML}{2563EB}
\definecolor{smlgreen}{HTML}{059669}
\definecolor{dncpurple}{HTML}{7C3AED}
\definecolor{problemred}{HTML}{DC2626}
\definecolor{lightblue}{HTML}{DBEAFE}
\definecolor{lightgreen}{HTML}{D1FAE5}
\definecolor{lightpurple}{HTML}{EDE9FE}
\definecolor{lightred}{HTML}{FEE2E2}
\definecolor{tablegray}{HTML}{F3F4F6}
\definecolor{ourscolor}{HTML}{EFF6FF}
\definecolor{reportedcolor}{gray}{0.93}

\definecolor{cBlue}{HTML}{2563EB}
\definecolor{cRed}{HTML}{DC2626}
\definecolor{cGreen}{HTML}{16A34A}
\definecolor{cOrange}{HTML}{EA580C}
\definecolor{cPurple}{HTML}{7C3AED}
\definecolor{cTeal}{HTML}{0D9488}
\definecolor{cGray}{HTML}{64748B}
\definecolor{cDark}{HTML}{1E293B}

\definecolor{deltagreen}{HTML}{16A34A}
\definecolor{deltared}{HTML}{DC2626}

\definecolor{lighttext}{gray}{0.55}

\newcommand{\sml}{\textsc{SML}}
\newcommand{\dps}{\textsc{DPS}}
\newcommand{\ours}{\textsc{DACA-GRPO}}

\title{\ours{}: Denoising-Aware Credit Assignment for Reinforcement Learning in Diffusion Language Models}

\author[1,2,\dagger,\ddagger]{Amin Karimi Monsefi}
\author[2,\dagger]{Dominic Culver}
\author[2]{Nikhil Bhendawade}
\author[2]{Lokesh Boominathan}
\author[2]{Manuel R. Ciosici}
\author[2]{Yizhe Zhang}
\author[2]{Irina Belousova}
\affiliation{$^{1}$The Ohio State University \quad $^{2}$Apple}

\abstract{

Diffusion large language models are a compelling alternative to autoregressive models, yet existing RL methods for diffusion treat all denoising steps as equally important and rely on biased, high-variance likelihood estimates. We identify two fundamental weaknesses: the absence of \emph{temporal credit assignment} across the denoising trajectory, and the systematic bias of mean-field likelihood estimates used for policy optimization. To address these, we propose Denoising-Aware Credit Assignment for GRPO (\ours{}), a lightweight, plug-and-play enhancement for any GRPO-style trainer. \ours{} introduces two complementary mechanisms: \emph{Denoising Progress Scores}, which extract per-token importance weights from intermediate predictions at no additional forward cost, and \emph{Stratified Masking Likelihood}, which partitions token positions into strata so that each token is predicted with most of the sequence as context, reducing the mean-field bias. Applied on top of three GRPO base methods, \ours{} achieves consistent improvements across seven benchmarks spanning mathematical reasoning, code generation, constraint satisfaction, and constrained generation, with gains of up to 5.6pp on math reasoning, 7.4pp on code generation, 36.3pp on constraint satisfaction, and 5.9pp on JSON schema adherence.

}

\metadata[Correspondence]{\sffamily
  Amin Karimi Monsefi: \url{karimimonsefi.1@osu.edu};
  Dominic Culver: \url{dominic_culver@apple.com};
  Irina Belousova: \url{ibelousova@apple.com}}
\date{\sffamily\today}

\begin{document}
\maketitle

\applefootnote{\textcolor{textgray}{\sffamily%
  $^{\dagger}$Equal contribution.\\
  $^{\ddagger}$The Ohio State University. Work done during the internship at Apple.}}


\section{Introduction}
\label{sec:intro}

Diffusion large language models (dLLMs) are an alternative paradigm to autoregressive models, supporting high-throughput, any-order parallel text generation. They operate by iteratively denoising corrupted sequences, starting from a sequence of mask or uniform tokens~\citep{mdlm,llada, monsefi2025fs,khanna2025mercury, ye2025dream}.

Group Relative Policy Optimization~\citep{guo2025deepseek,yu2025dapo,liu2025understanding} (GRPO), now a standard post-training method for autoregressive LLMs, has recently been adapted to discrete diffusion models~\citep{d1, wd1, rojas2025improving, gong2025diffucoder}. Unlike autoregressive LLMs, dLLMs cannot directly compute the completion log-likelihood $\log p_{\theta}(o | q)$, so it must be approximated or replaced with an evidence lower bound (ELBO).

While these methods have seen significant success, they ignore the \emph{denoising trajectory} itself. During generation, dLLMs 
compute distributions over every masked position at each step but unmask only a few; the remaining predictions---encoding the model's evolving understanding---are discarded. Not all steps contribute equally: some involve decisive structural commitments, while others are routine fill-in. Yet current methods assign equal credit to every token regardless of when or how confidently it was generated. Moreover, the likelihood estimates suffer from mean-field bias---predicting all tokens from zero inter-token context---corrupting the policy gradient. We make the following contributions:

\begin{enumerate}
    \item We \textbf{identify two fundamental weaknesses} in existing dLLM RL methods: the absence of temporal credit assignment across denoising steps, and the systematic bias of mean-field likelihood estimates used for policy optimization.

    \item \textbf{Denoising Progress Scores (\dps{})} track the model's belief evolution across denoising steps, converting these signals into per-token importance weights that modulate the RL loss. \dps{} requires \emph{no additional forward passes}---it repurposes logits that the model would otherwise discard.

    \item To mitigate the likelihood bias, we introduce \textbf{Stratified Masking Likelihood (\sml{})}, a position-stratified estimator that partitions output tokens into $K$ strata and evaluates each token with $(K{-}1)/K$ of the sequence as context. Intuitively, this should reduce the noise introduced by using the mean-field approximation by ensuring that token interdependency is captured amongst the strata. 
    

    \item \textbf{\ours{}} combines \dps{} and \sml{} into a unified, plug-and-play framework that is agnostic to the underlying GRPO variant.
\end{enumerate}

We evaluate \ours{} on top of three popular RL methods---Diffu-GRPO~\citep{d1}, wd1~\citep{wd1}, and GDPO~\citep{rojas2025improving}---across seven benchmarks spanning mathematical reasoning (MATH-500, GSM8K), code generation (MBPP, HumanEval), constraint satisfaction (Countdown, Sudoku), and constrained generation (JSON). \ours{} delivers consistent improvements across benchmarks and base methods, with gains of up to 5.6pp on math reasoning, 7.4pp on code generation, 36.3pp on constraint satisfaction, and 5.9pp on JSON schema adherence. The two components are complementary: \dps{} is most impactful on structured constraint tasks and short generation lengths, while \sml{} provides the largest gains on long-form math and code tasks.

\begin{figure}[t]
	\centering
	\includegraphics[width=\textwidth]{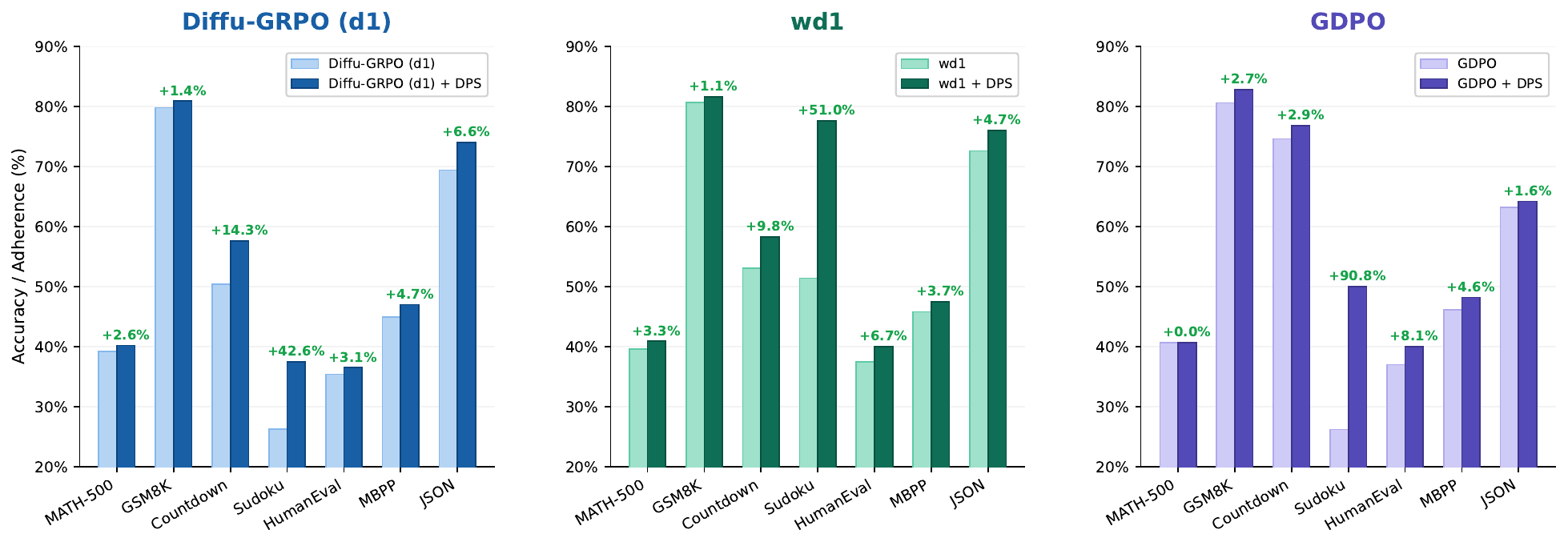}
	\caption{\textbf{\dps{} improves all three base methods across all benchmarks.} Best accuracy (\%) across generation lengths \{128, 256, 512\}. Light bars: reproduced baselines. Dark bars: with \dps{} applied. Green percentages show relative improvement over the baseline. \dps{} delivers consistent gains across mathematical reasoning, code generation, constraint satisfaction, and constrained generation tasks, with the largest relative gains on Sudoku---up to $+51\%$ for wd1 and $+90\%$ for GDPO.}
	\label{fig:dps-results}
\end{figure}
\section{Related work}
\label{sec:related}

Diffusion language models like MDLM~\citep{mdlm} and LLaDA~\citep{llada} established masked diffusion as a viable text generation paradigm, and subsequent work has applied direct preference optimization to this setting~\citep{ddpo,xie2025dream,lou2023discrete}. Our work is orthogonal to the diffusion architecture; we focus on the RL training signal.

Reinforcement learning has been widely applied to autoregressive LLMs through methods like Reinforcement Learning from Human Feedback~(\citet{rlhf}) and GRPO~\citep{grpo}. Recent work extends GRPO to diffusion models: Diffu-GRPO~\citep{d1} adapts PPO-clipped objectives, wd1~\citep{wd1} proposes a ratio-free alternative, GDPO~\citep{rojas2025improving} formulates diffusion-specific policy optimization, and DiffuCoder~\citep{gong2025diffucoder} introduces coupled sampling for variance reduction. All four treat the denoising trajectory as a black box, assigning uniform credit across steps. \ours{} is the first to exploit trajectory-level signals for credit assignment.

Temporal credit assignment is a classical challenge in reinforcement learning~\citep{sutton_barto}. In the autoregressive setting, per-token reward models~\citep{prm, zhu2025surprising} and process reward models~\citep{process_reward} provide step-level supervision but require training additional reward models or collecting extra annotations. \dps{} addresses the same challenge by exploiting a structural property of diffusion models: intermediate predictions are already computed during generation but usually discarded. Concurrently, DyLLM~\citep{leeDyLLMEfficientDiffusion2026} observes that not all token positions contribute equally across denoising steps and identifies \emph{salient} positions via attention-context similarity to accelerate inference. Both methods mine signals from the denoising trajectory, but to different ends: DyLLM uses attention-based saliency to accelerate inference, while \dps{} uses token-level prediction confidence to improve \emph{training} via credit assignment.

\sml{} applies a different principle to masked-token likelihood estimation: instead of stratifying the masking \emph{ratio} (as in DiffuCoder's antithetic timesteps~\citep{gong2025diffucoder}) or using deterministic quadrature over time (as in GDPO's SDMC~\citep{rojas2025improving}), we stratify token \emph{positions}---masking one stratum at a time to provide partial context, targeting mean-field estimation error rather than sampling \emph{variance}.
\section{Background}
\label{sec:background}

Masked diffusion language models~\citep{mdlm,llada} define a forward process that progressively masks tokens and a reverse process that denoises them. Given a clean sequence $\mathbf{x}_0 = (x_1, \ldots, x_L)$, the forward process produces a noisy sequence $\mathbf{x}_t$ at timestep $t \in [0, 1]$ by independently replacing each token with \texttt{[MASK]} with probability $t$. The reverse (generative) process starts with a fully masked sequence $\mathbf{x}_1$ and iteratively predicts and reveals tokens until reaching $\mathbf{x}_0$. At each denoising step $t$, the model $p_\theta$ predicts a distribution $p_\theta(x_i \mid \mathbf{x}_t)$ over all vocabulary tokens for every masked position $i \in \mathcal{M}_t$, where $\mathcal{M}_t = \{i : x_{t,i} = \texttt{[MASK]}\}$. A fixed number $k$ of masked positions are then unmasked by selecting positions and tokens with the highest model confidence. In particular, predictions are made for \emph{all} masked positions but most are discarded. These ``wasted'' predictions form the basis of our \dps{} mechanism.

GRPO for diffusion language models~\citep{grpo} generates a group of $G$ candidate responses $\{y^{(g)}\}_{g=1}^G$ for each prompt, computes rewards $\{r^{(g)}\}$, and defines group-relative advantages using the group mean $\mu_r$ and standard deviation $\sigma_r$:
\begin{equation}
    A^{(g)} = \frac{r^{(g)} - \mu_r}{\sigma_r}.
    \label{eq:advantage}
\end{equation}

\textbf{Diffu-GRPO}~\citep{d1} adapts GRPO to dLLMs using a per-token PPO-clipped surrogate objective with per-token likelihood ratio $\rho_i(\theta) = p_\theta(x_i \mid \mathbf{m}) / p_{\theta_{\text{old}}}(x_i \mid \mathbf{m})$, computed with all completion tokens masked ($\mathbf{m}$):
\begin{equation}
    \mathcal{L}_{\text{d1}} = -\mathbb{E}\left[\min\left(\rho_i(\theta) A, \; \text{clip}(\rho_i(\theta), 1-\epsilon, 1+\epsilon) A\right)\right].
    \label{eq:d1}
\end{equation}

\textbf{wd1}~\citep{wd1} proposes a ratio-free alternative with explicit positive and negative sample reinforcement, using normalized advantage weights $\hat{A}^{(g)}_+ = \text{softmax}(A^{(g)})$ and $\hat{A}^{(g)}_- = \text{softmax}(-A^{(g)})$:
\begin{equation}
    \mathcal{L}_{\text{wd1}} = \underbrace{-\sum_{g: A^{(g)} > 0} \hat{A}^{(g)}_+ \log p_\theta(y^{(g)})}_{\text{PSR: amplify correct}} \; \underbrace{+ \sum_{g: A^{(g)} < 0} \hat{A}^{(g)}_- \log p_\theta(y^{(g)})}_{\text{NSR: suppress incorrect}}.
    \label{eq:wd1}
\end{equation}

\textbf{GDPO}~\citep{rojas2025improving} operates at the \emph{sequence level}, using the evidence lower bound (ELBO) as a surrogate for $\log p_\theta(y)$:
\begin{equation}
    \mathcal{L}_{\text{ELBO}}(y \mid q) = \mathbb{E}_{t \sim \mathcal{U}[0,1]} \, \mathbb{E}_{y_t \sim \pi_t(\cdot | y)} \left[\frac{1}{t} \sum_{i=1}^{L} \mathbf{1}[y_{t,i} = \texttt{M}] \log \pi_\theta(y_i \mid y_t, q)\right].
    \label{eq:elbo}
\end{equation}
To reduce variance, GDPO introduces Semi-deterministic Monte Carlo (SDMC): $N$ fixed quadrature points over $t$ with random mask sampling. The loss uses sequence-level importance ratios $r_g = \mathcal{L}_{\text{ELBO}}^{\theta}(y_g \mid q) \,/\, \mathcal{L}_{\text{ELBO}}^{\theta_{\text{old}}}(y_g \mid q)$ and unnormalized advantages $A_g = r_g - \text{mean}(r_1, \ldots, r_G)$:
\begin{equation}
    \mathcal{L}_{\text{GDPO}}(\theta) = -\frac{1}{G} \sum_{g=1}^{G} \frac{1}{|y_g|} \min\!\left(r_g \cdot A_g, \; \text{clip}(r_g, 1-\epsilon, 1+\epsilon) \cdot A_g\right) + \beta \, \text{KL}(\pi_\theta \| \pi_{\text{ref}}).
    \label{eq:gdpo}
\end{equation}

\paragraph{Masked-token log-likelihood.}
\label{sec:bg-loglik}
All three methods require estimating $\log p_\theta(y)$ for each completion. In dLLMs, this is typically approximated by sampling a random masking ratio $t \sim \mathcal{U}(0,1)$, constructing a masked version $\mathbf{x}_t$, and computing:
\begin{equation}
    \log p_\theta(y) \approx \frac{1}{|\mathcal{M}_t|} \sum_{i \in \mathcal{M}_t} \log p_\theta(x_i \mid \mathbf{x}_t).
    \label{eq:loglik}
\end{equation}
Here $p_\theta(x_i \mid \mathbf{x}_t)$ is obtained from the model's output logits at position $i$ via softmax---each per-token log-probability is exact; the approximation arises from the mean-field factorization into independent per-token terms.
d1 and wd1 use this single-sample estimate directly, incurring high variance and mean-field bias (all tokens masked, zero inter-token context). GDPO reduces variance via deterministic quadrature over $t$, but the per-point estimates remain subject to the same bias. \sml{} addresses the mean-field bias by predicting each token with $(K{-}1)/K$ of the sequence as context.

\section{Weaknesses of current approaches}
\label{sec:weaknesses}

\textbf{Weakness 1: No Temporal Credit Assignment}
\label{sec:w1}
In autoregressive models, the sequential generation process naturally induces a form of credit assignment: each token's probability is conditioned on all preceding tokens, and per-token rewards or advantages can be assigned based on position. Diffusion models lack this structure. All tokens are predicted simultaneously at each step, and the RL loss (\cref{eq:d1,eq:wd1}) assigns a single scalar advantage $A^{(g)}$ to the \emph{entire} completion. Denoising steps vary dramatically in importance: some involve decisive transitions where the model commits to key structural choices, while others are routine fill-in with minimal belief change. Standard methods assign equal weight to all steps and discard the intermediate predictions entirely. More specifically, uniform weighting under-credits \emph{decisive steps} that involve major belief changes---committing to key tokens such as operators in math or control flow in code---while over-crediting \emph{routine steps} with minimal belief change. It also over-punishes \emph{self-correcting steps} where the model doubts an incorrect answer, despite already moving toward the right solution. The result is inefficient gradient updates that fail to concentrate learning on the steps that matter most.

 \textbf{Weakness 2: Biased and High-Variance Likelihood Estimates}
 \label{sec:w2}
 All three base methods require estimating $\log p_\theta(y)$ for each completion. The standard approach (\cref{eq:loglik}) masks \emph{all} completion tokens and predicts them from zero context, treating tokens as independent. This introduces \emph{systematic bias}: predicting without any context is fundamentally harder, adding errors to the log likelihood estimates. It also introduces \emph{high variance}: a single random masking pattern may over- or under-represent the typical difficulty of predicting $y$, producing noisy gradients.
\section{Method}
\label{sec:method}

We present \ours{}, which addresses both weaknesses through two complementary mechanisms: \emph{Denoising Progress Scores} and \emph{Stratified Masking Likelihood}.

\subsection{Denoising progress scores (\dps{})}
\label{sec:dps}

At each denoising step, the model predicts token distributions for all masked positions, but only unmasks a few. The predictions at still-masked positions reveal how well the model ``understands'' the emerging sequence---and how much that understanding changes between steps provides a natural measure of step importance. 

Formally, let $\mathbf{x}_0, \mathbf{x}_1, \ldots, \mathbf{x}_T$ denote the denoising trajectory and $\mathcal{M}_T \subset \mathcal{M}_{T-1} \subset \cdots \subset \mathcal{M}_0$ the family of masked positions ($\mathbf{x}_0$ fully masked, $\mathbf{x}_T$ fully revealed). We define $S(k, j)$ as the average log-probability under the model at step $k$ over the positions in $\mathcal{M}_j$:
\begin{equation}
    S(k, j) = \frac{1}{|\mathcal{M}_j|} \sum_{i \in \mathcal{M}_j} \log p_\theta(x^*_i \mid \mathbf{x}_k),
    \label{eq:score}
\end{equation}

where $x^*_i$ is the final token at position $i$ in the model's own generated completion, i.e., the token the model itself selected during the rollout, not an external label. The \emph{denoising progress delta} at step $k$ measures the belief improvement when transitioning from step $k$ to $k+1$:
\begin{equation}
	\delta_k := S(k{+}1,\, k{+}1) - S(k,\, k{+}1)
	\label{eq:delta}
\end{equation}

\begin{figure}[t]
\centering
\begin{tikzpicture}[
    scale=0.82, transform shape,
    maskbox/.style={rectangle, minimum width=0.4cm, minimum height=0.4cm, inner sep=0pt, font=\scriptsize},
    revealed/.style={maskbox, fill=dpsblue!30, draw=dpsblue!50},
    masked/.style={maskbox, fill=gray!20, draw=gray!40},
    evalbox/.style={maskbox, fill=smlgreen!20, draw=smlgreen!50, font=\scriptsize},
]

\node[font=\small\bfseries, anchor=west] at (-0.5, 4.0) {\dps{} Score Computation (\cref{eq:score})};

\node[font=\small, anchor=east] at (-0.3, 3.0) {Step $k$:};
\node[font=\small, anchor=east] at (-0.3, 2.0) {Step $k{+}1$:};
\node[font=\small, anchor=east] at (-0.3, 1.1) {Evaluation:};

\foreach \i in {0,...,11} {
    \pgfmathparse{(\i < 3) ? 1 : 0}
    \ifnum\pgfmathresult=1
        \node[revealed] at ({\i*0.55}, 3.0) {};
    \else
        \node[masked] at ({\i*0.55}, 3.0) {\textcolor{gray!60}{M}};
    \fi
}

\foreach \i in {0,...,11} {
    \pgfmathparse{(\i < 5) ? 1 : 0}
    \ifnum\pgfmathresult=1
        \node[revealed] at ({\i*0.55}, 2.0) {};
    \else
        \node[masked] at ({\i*0.55}, 2.0) {\textcolor{gray!60}{M}};
    \fi
}

\foreach \i in {0,...,11} {
    \pgfmathparse{(\i >= 5) ? 1 : 0}
    \ifnum\pgfmathresult=1
        \node[evalbox] at ({\i*0.55}, 1.1) {\checkmark};
    \else
        \node[maskbox, fill=white, draw=gray!20] at ({\i*0.55}, 1.1) {--};
    \fi
}

\draw[decorate, decoration={brace, amplitude=4pt}, dpsblue!60] (2.6, 2.35) -- (6.22, 2.35);
\node[font=\scriptsize, dpsblue!80!black] at (4.25, 2.65) {newly revealed at $k{+}1$};

\draw[decorate, decoration={brace, amplitude=4pt, mirror}, smlgreen!60] (2.75, 0.75) -- (6.02, 0.75);
\node[font=\scriptsize, smlgreen!80!black] at (4.38, 0.4) {evaluate belief at both steps};

\node[font=\small, anchor=west] at (7.2, 2.2) {$\delta_k = S(k{+}1, k{+}1) - S(k, k{+}1)$};
\node[font=\footnotesize, color=gray, anchor=west] at (7.2, 1.6) {Measures belief improvement from step $k$ to $k{+}1$};
\node[font=\footnotesize, color=dpsblue, anchor=west] at (7.2, 1.1) {\textbf{Cost: zero} --- logits already computed};

\node[revealed, label={[font=\scriptsize]right:revealed}] at (7.5, 3.5) {};
\node[masked, label={[font=\scriptsize]right:masked}] at (9.5, 3.5) {\textcolor{gray!60}{M}};
\node[evalbox, label={[font=\scriptsize]right:evaluated}] at (11.5, 3.5) {\checkmark};

\end{tikzpicture}
\caption{\textbf{\dps{} score computation.} We quantify the change in model beliefs by computing the difference $\delta_k$ in model predictions on positions masked in the current and the following step. \dps{} uses the logits computed naturally during generation and requires no extra forward passes.}
\label{fig:dps-computation}
\end{figure}
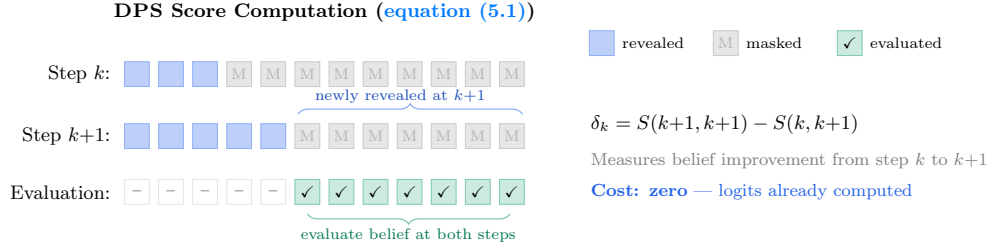

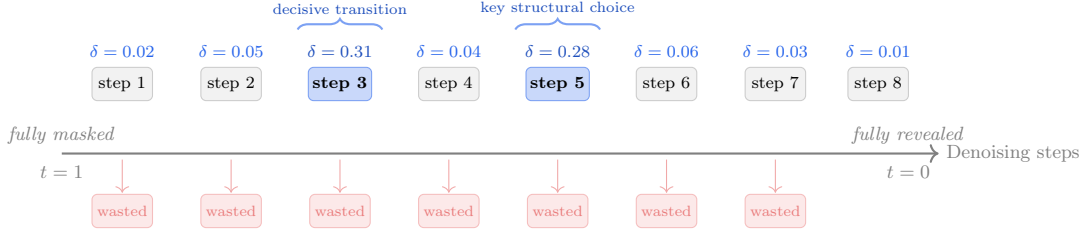
\begin{figure}[t]
	\centering
	\begin{tikzpicture}[
		scale=0.8, transform shape,
		mystep/.style={
			rectangle,
			rounded corners=2pt,
			minimum width=1.0cm,
			minimum height=0.55cm,
			font=\footnotesize,
			draw=gray!50,
			inner sep=2pt
		},
		important/.style={
			mystep,
			fill=dpsblue!25,
			draw=dpsblue!60,
			font=\footnotesize\bfseries
		},
		routine/.style={
			mystep,
			fill=gray!10,
			draw=gray!40
		},
		discarded/.style={
			mystep,
			fill=problemred!10,
			draw=problemred!30,
			text=problemred!60
		}
		]
		
		\draw[->, thick, gray] (0,0) -- (14.5,0) node[right, font=\small] {Denoising steps};
		\node[font=\small, gray] at (0, -0.3) {$t=1$};
		\node[font=\small, gray] at (14, -0.3) {$t=0$};
		\node[font=\small, gray] at (0, 0.3) {\textit{fully masked}};
		\node[font=\small, gray] at (14, 0.3) {\textit{fully revealed}};
		
		\node[routine]   (s1) at (1,   1.15) {step 1};
		\node[routine]   (s2) at (2.8, 1.15) {step 2};
		\node[important] (s3) at (4.6, 1.15) {step 3};
		\node[routine]   (s4) at (6.4, 1.15) {step 4};
		\node[important] (s5) at (8.2, 1.15) {step 5};
		\node[routine]   (s6) at (10,  1.15) {step 6};
		\node[routine]   (s7) at (11.8,1.15) {step 7};
		\node[routine]   (s8) at (13.5,1.15) {step 8};
		
		\node[font=\footnotesize, dpsblue] at (1,   1.7) {$\delta=0.02$};
		\node[font=\footnotesize, dpsblue] at (2.8, 1.7) {$\delta=0.05$};
		\node[font=\footnotesize\bfseries, dpsblue!80!black] at (4.6, 1.7) {$\delta=0.31$};
		\node[font=\footnotesize, dpsblue] at (6.4, 1.7) {$\delta=0.04$};
		\node[font=\footnotesize\bfseries, dpsblue!80!black] at (8.2, 1.7) {$\delta=0.28$};
		\node[font=\footnotesize, dpsblue] at (10,  1.7) {$\delta=0.06$};
		\node[font=\footnotesize, dpsblue] at (11.8,1.7) {$\delta=0.03$};
		\node[font=\footnotesize, dpsblue] at (13.5,1.7) {$\delta=0.01$};
		
		\foreach \x in {1, 2.8, 4.6, 6.4, 8.2, 10, 11.8} {
			\node[discarded, font=\scriptsize] at (\x, -0.95) {wasted};
			\draw[->, problemred!40, thin] (\x, -0.1) -- (\x, -0.65);
		}
		
		\draw[decorate, decoration={brace, amplitude=4pt, mirror}, dpsblue!60, thick]
		(5.3, 2.05) -- (3.9, 2.05);
		\node[font=\scriptsize, dpsblue!80!black] at (4.6, 2.4) {decisive transition};
		
		\draw[decorate, decoration={brace, amplitude=4pt, mirror}, dpsblue!60, thick]
		(8.9, 2.05) -- (7.5, 2.05);
		\node[font=\scriptsize, dpsblue!80!black] at (8.2, 2.4) {key structural choice};
		
	\end{tikzpicture}
	\caption{\textbf{Illustration of Weakness 1.} During generation, dLLMs perform many denoising steps of varying importance. Steps 3 and 5 involve decisive transitions (large $\delta$), while others are routine. Standard methods assign equal credit to all steps and discard intermediate predictions.}
	\label{fig:credit-assignment}
\end{figure}

Both scores are evaluated on the same set of positions $\mathcal{M}_{k+1}$ (i.e., those still masked at the next step), so the difference isolates the effect of the newly revealed tokens (\cref{fig:dps-computation}). A large positive $\delta$ indicates that revealing tokens at step $k{+}1$ substantially improved the model's understanding (i.e., an important denoising step). A small or negative $\delta$ indicates minimal or confused progress (\cref{fig:credit-assignment}).

\textbf{\dps{} requires no additional forward passes}. At each denoising step, the model already computes logits for all masked positions; \dps{} reuses the log-probabilities at still-masked positions, values that would otherwise be discarded. The only overhead is one scalar per recorded step per sample (<1\% wall-clock; \cref{app:compute-overhead}).

Raw deltas vary in scale across denoising steps due to the non-stationary nature of the denoising process (early steps have little context and small deltas; late steps have rich context and larger deltas). Therefore, we apply \emph{per-step} z-score normalization across the batch:
\begin{equation}
    \bar{\delta}_k = \frac{\delta_k - \mu_k}{\sigma_k + \epsilon},
    \label{eq:normalize}
\end{equation}
where $\mu_k$ and $\sigma_k$ are the batch mean and standard deviation of $\delta_k$. This ensures fair credit assignment across the entire denoising timeline: an ``impressive'' early-step delta and an ``impressive'' late-step delta receive comparable normalized values, despite different magnitudes. We compared four alternatives and found that per-step exceeds the baseline across all stride configurations (\Cref{app:ablation-normalization}).

\textbf{Per-token weight modulation.} Each token in the final sequence is ``born'' at the denoising step where it was unmasked. We track birth step and define the \emph{\dps{}-modulated per-token advantage}:
\begin{equation}
    \tilde{A}_i^{(g)} = A^{(g)} \cdot \bigl(1 + \lambda \cdot \bar{\delta}_{\text{birth}(i)}\bigr),
    \label{eq:weight}
\end{equation}
where $\lambda > 0$ controls modulation strength. The factor $\omega_i = 1 + \lambda \cdot \bar{\delta}_{\text{birth}(i)}$ amplifies or attenuates the learning signal based on step importance. For negative-advantage samples, the sign of $\bar{\delta}$ provides directional credit: positive values increase punishment (confident progress toward the wrong answer), negative ones decrease it (hesitation or self-correction).

\dps{} implementation requires two design choices:
(1) tokens unmasked at the final recorded denoising step have no following step to compute a delta against, so we extrapolate $\delta_T = \delta_{T-1}$; (2) we record snapshots with a stride $s$, trading credit granularity for signal-to-noise. Both choices are robust: extrapolation outperforms four alternatives, and every stride $s \in \{1, 2, 4, 8, 16, 32\}$ exceeds the baseline under our default settings. Ablations and detailed discussion in \cref{sec:ablation-dps-stride,app:method-details}.

\subsection{Stratified masking likelihood (\sml{})}
\label{sec:sml}

Log-likelihood estimates used by all three base methods mask \emph{all} completion tokens and predict them from zero inter-token context. This mean-field approximation introduces error (\cref{eq:d1,eq:gdpo,eq:wd1}). To mitigate this, we propose a stratified estimator that uses $K$ forward passes. We partition the $N$ output token positions $o_1, \ldots, o_N$ of completion $y$ into $K$ disjoint strata $\mathcal{S}_1, \ldots, \mathcal{S}_K$ of roughly equal size. For each stratum $\mathcal{S}_k$, we mask \emph{only} the positions in $\mathcal{S}_k$ while keeping all other output tokens visible, and compute per-token log-probability for each masked position $n \in \mathcal{S}_k$. Summing the token-level log-probabilities across all strata gives the sequence-level \sml{} estimate:
\begin{equation}
    \log p_\theta^{\text{SML}}(o | q) = \sum_{k=1}^{K} \sum_{n \in \mathcal{S}_k} \log p_\theta(o_n \mid o_{\setminus \mathcal{S}_k}, q),
    \label{eq:sml}
\end{equation}

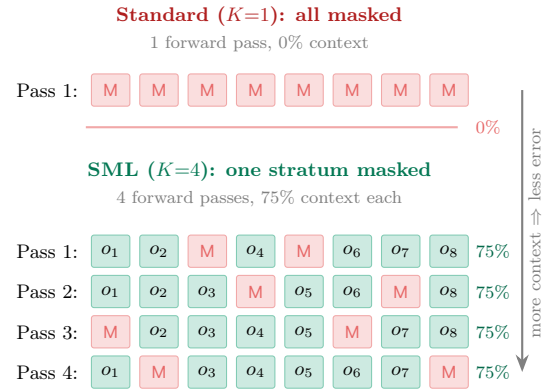
\begin{wrapfigure}{r}{0.48\textwidth}
\vspace{-12pt}
\centering
\begin{tikzpicture}[
    scale=0.82, transform shape,
    tok/.style={rectangle, minimum width=0.62cm, minimum height=0.52cm, inner sep=0pt, font=\small, rounded corners=1pt},
    vis/.style={tok, fill=smlgreen!20, draw=smlgreen!50},
    msk/.style={tok, fill=problemred!15, draw=problemred!40, text=problemred!70},
    lbl/.style={font=\small, anchor=east},
]

\node[font=\small\bfseries, color=problemred!80!black] at (2.8, 3.5) {Standard ($K{=}1$): all masked};
\node[font=\footnotesize, color=gray] at (2.8, 3.05) {1 forward pass, 0\% context};

\node[lbl] at (0.0, 2.3) {Pass 1:};
\node[msk] at (0.40, 2.3) {\textsf{M}};
\node[msk] at (1.18, 2.3) {\textsf{M}};
\node[msk] at (1.96, 2.3) {\textsf{M}};
\node[msk] at (2.74, 2.3) {\textsf{M}};
\node[msk] at (3.52, 2.3) {\textsf{M}};
\node[msk] at (4.30, 2.3) {\textsf{M}};
\node[msk] at (5.08, 2.3) {\textsf{M}};
\node[msk] at (5.86, 2.3) {\textsf{M}};

\draw[problemred!40, thick] (0, 1.7) -- (6.0, 1.7);
\node[font=\footnotesize, problemred!70, anchor=west] at (6.17, 1.7) {0\%};

\node[font=\small\bfseries, color=smlgreen!80!black] at (2.8, 1.0) {\sml{} ($K{=}4$): one stratum masked};
\node[font=\footnotesize, color=gray] at (2.8, 0.55) {4 forward passes, 75\% context each};

\node[lbl] at (0.0, -0.3) {Pass 1:};
\node[vis] at (0.40, -0.3) {$o_1$};
\node[vis] at (1.18, -0.3) {$o_2$};
\node[msk] at (1.96, -0.3) {\textsf{M}};
\node[vis] at (2.74, -0.3) {$o_4$};
\node[msk] at (3.52, -0.3) {\textsf{M}};
\node[vis] at (4.30, -0.3) {$o_6$};
\node[vis] at (5.08, -0.3) {$o_7$};
\node[vis] at (5.86, -0.3) {$o_8$};
\node[font=\footnotesize, smlgreen!70!black, anchor=west] at (6.17, -0.3) {75\%};

\node[lbl] at (0.0, -0.95) {Pass 2:};
\node[vis] at (0.40, -0.95) {$o_1$};
\node[vis] at (1.18, -0.95) {$o_2$};
\node[vis] at (1.96, -0.95) {$o_3$};
\node[msk] at (2.74, -0.95) {\textsf{M}};
\node[vis] at (3.52, -0.95) {$o_5$};
\node[vis] at (4.30, -0.95) {$o_6$};
\node[msk] at (5.08, -0.95) {\textsf{M}};
\node[vis] at (5.86, -0.95) {$o_8$};
\node[font=\footnotesize, smlgreen!70!black, anchor=west] at (6.17, -0.95) {75\%};

\node[lbl] at (0.0, -1.6) {Pass 3:};
\node[msk] at (0.40, -1.6) {\textsf{M}};
\node[vis] at (1.18, -1.6) {$o_2$};
\node[vis] at (1.96, -1.6) {$o_3$};
\node[vis] at (2.74, -1.6) {$o_4$};
\node[vis] at (3.52, -1.6) {$o_5$};
\node[msk] at (4.30, -1.6) {\textsf{M}};
\node[vis] at (5.08, -1.6) {$o_7$};
\node[vis] at (5.86, -1.6) {$o_8$};
\node[font=\footnotesize, smlgreen!70!black, anchor=west] at (6.17, -1.6) {75\%};

\node[lbl] at (0.0, -2.25) {Pass 4:};
\node[vis] at (0.40, -2.25) {$o_1$};
\node[msk] at (1.18, -2.25) {\textsf{M}};
\node[vis] at (1.96, -2.25) {$o_3$};
\node[vis] at (2.74, -2.25) {$o_4$};
\node[vis] at (3.52, -2.25) {$o_5$};
\node[vis] at (4.30, -2.25) {$o_6$};
\node[vis] at (5.08, -2.25) {$o_7$};
\node[msk] at (5.86, -2.25) {\textsf{M}};
\node[font=\footnotesize, smlgreen!70!black, anchor=west] at (6.17, -2.25) {75\%};

\draw[thick, -{Stealth}, gray] (7.04, 2.3) -- (7.04, -2.25);
\node[font=\footnotesize, gray, align=center, rotate=90] at (7.24, 0.0) {more context $\Rightarrow$ less error};

\end{tikzpicture}
\vspace{-6pt}
\caption{\textbf{Stratified Masking Likelihood.} Standard estimation ($K{=}1$) masks all tokens with zero context. \sml{} ($K{=}4$) masks one stratum per pass, giving each token 75\% context.}
\label{fig:sml}
\vspace{-8pt}
\end{wrapfigure}

where $o_{\setminus \mathcal{S}_k}$ denotes the output sequence with only stratum $\mathcal{S}_k$ replaced by mask tokens. Each token is predicted with $(K{-}1)/K$ of the sequence as visible context, forming a tunable spectrum: $K{=}1$ places all tokens in a single stratum, recovering the fully-masked estimate of \cref{eq:loglik}; $K{=}N$ gives each token its own stratum, recovering the pseudo-log-likelihood where every token sees all others (\cref{fig:sml}). We hypothesize that this method of approximating the log probabilities reduces the estimation errors in mean-field based methods such as d1 and we verify this hypothesis empirically in \Cref{sec:experiments}. \sml{} targets the \emph{bias} component of mean-field error---complementary to variance-reduction methods like GDPO's SDMC.

\subsection{\sml{} integration}
\label{sec:sml-integration}

Integrating \sml{} into the training loss depends on whether the base method uses a clipped importance ratio.

\paragraph{Ratio-Free Methods}
\label{sec:sml-wd1}
like wd1 optimize log-probabilities directly without importance ratios or trust regions. Adding an \sml{}-based likelihood term is therefore straightforward---both losses operate on $\log p_\theta$ and their gradients compose without conflict:
\begin{equation}
    \mathcal{L}_{\text{DACA-wd1}} = \mathcal{L}_{\text{wd1}} - \frac{\eta}{G} \sum_{g=1}^{G} \log p_\theta^{\text{SML}}(y^{(g)}),
    \label{eq:dnc-wd1}
\end{equation}
where $\eta$ controls the SML loss strength. The base loss handles reward-driven credit assignment (amplifying correct answers, suppressing incorrect ones), while the \sml{} term provides a complementary signal that improves token-prediction quality under richer context. 

\paragraph{Ratio-Based Methods}
\label{sec:sml-ratio}
like d1 and GDPO (\cref{eq:d1,eq:gdpo}) cannot accept an additive \sml{} loss as it destabilizes training: its gradient bypasses the PPO clip, resulting in training collapse. Instead, we incorporate \sml{} \emph{inside} the importance ratio at token level. For each output position $i$, let $k(i)$ denote the stratum containing position $i$. We form an enriched per-token log-probability by averaging two views---the fully-masked prediction and the stratum prediction:
\begin{equation}
    \hat{\ell}_{\theta,i} = \frac{1}{2}\!\left[\log p_\theta(o_i \mid \mathbf{m}, q) + \log p_\theta(o_i \mid o_{\setminus \mathcal{S}_{k(i)}}, q)\right],
    \label{eq:enriched-ll}
\end{equation}
where $\mathbf{m}$ denotes the fully-masked completion (the standard estimate from \cref{eq:loglik}) and $o_{\setminus \mathcal{S}_{k(i)}}$ denotes the completion with only stratum $\mathcal{S}_{k(i)}$ masked. This yields a per-token importance ratio:
\begin{equation}
    \rho_i^{\text{SML}} := \exp\!\left(\hat{\ell}_{\theta,i} - \hat{\ell}_{\theta_{\text{old}},i}\right),
    \label{eq:ratio-sml}
\end{equation}
which replaces the standard per-token ratio in the clipped surrogate. Each token's enriched log-probability blends the fully-masked view (matching the generation regime) with its stratified view (reducing estimation errors). The trust region is preserved by construction ($\rho_i^{\text{SML}}(\theta_{\text{old}}) = 1$); $\hat{\ell}_{\theta_{\text{old}},i}$ is computed once and cached. No hyperparameters beyond $K$ are introduced. \sml{} adds $K$ forward passes during loss computation---negligible relative to the generation phase that dominates each training step (see \cref{app:compute-overhead}).

\begin{algorithm}[t]
\caption{\ours{}}
\label{alg:dnc}
\begin{algorithmic}[1]
\REQUIRE Policy $\pi_\theta$, reward $R$, stride $s$, modulation strength $\lambda$, strata $K$
\FOR{each training step}
    \STATE Generate $G$ completions; cache logits at still-masked positions every $s$ steps
    \STATE Compute rewards and group-relative advantages $A^{(g)}$
    \STATE \dps{}: compute progress deltas $\delta_k$ from cached logits \hfill {\scriptsize \cref{eq:score,eq:delta}}
    \STATE \dps{}: per-token advantages $\tilde{A}_i^{(g)} \leftarrow A^{(g)} \cdot (1 + \lambda \cdot \bar{\delta}_{\text{birth}(i)})$ \hfill {\scriptsize \cref{eq:normalize,eq:weight}}
    \FOR{each inner iteration}
        \STATE \sml{}: compute $K$-stratum likelihood $\log p_\theta^{\text{SML}}(y)$ \hfill {\scriptsize \cref{eq:sml}}
        \STATE Compute $\mathcal{L}_{\text{DACA}}$ using $\tilde{A}_i^{(g)}$ and \sml{} likelihood \hfill {\scriptsize \cref{eq:dnc-wd1-full} or \cref{eq:dnc-d1-full}}
        \STATE Update $\theta$ via gradient descent
    \ENDFOR
\ENDFOR
\end{algorithmic}
\end{algorithm}

\subsection{\ours{}: combining \dps{} and \sml{}}
\label{sec:dnc}

\dps{} and \sml{} can be applied separately or jointly, as they modify two independent inputs to the loss: \dps{} modulates the per-token advantage while \sml{} improves the likelihood estimate. For ratio-free methods, the \textbf{combined loss} becomes:

\begin{equation}
    \mathcal{L}_{\text{DACA}} = \mathcal{L}_{\text{wd1}}\!\left(\tilde{A}_i^{(g)}\right) - \frac{\eta}{G}\sum_{g=1}^{G} \log p_\theta^{\text{SML}}(y^{(g)}),
    \label{eq:dnc-wd1-full}
\end{equation}
where the first term is the wd1 loss, but with \dps{} per-token advantages, and the second is the \sml{} loss. For ratio-based methods, \sml{} enters inside the importance ratio and \dps{} modulates the per-token advantages:
\begin{equation}
    \mathcal{L}_{\text{DACA}} = -\mathbb{E}\left[\min\left(\rho_i^{\text{SML}} \cdot \tilde{A}_i^{(g)}, \; \text{clip}(\rho_i^{\text{SML}}, 1{-}\epsilon, 1{+}\epsilon) \cdot \tilde{A}_i^{(g)}\right)\right].
    \label{eq:dnc-d1-full}
\end{equation}
\cref{alg:dnc} gives the complete procedure. The key structural property is that \dps{} weights are computed once per training step during generation (at zero additional cost) and reused across all inner iterations, while \sml{} forward passes occur within each inner iteration during loss computation.

\section{Experiments}
\label{sec:experiments}
 
\paragraph{Setup.}
\label{sec:setup}
 We use LLaDA-8B-Instruct~\citep{llada} as the base dLLM, consistent with prior work~\citep{d1,wd1}. We evaluate \ours{} on top of three GRPO variants: Diffu-GRPO~\citep{d1} (PPO-clipped surrogate with importance ratios), wd1~\citep{wd1} (ratio-free weighted log-likelihood), and GDPO~\citep{rojas2025improving} (token-level diffusion policy optimization).
 
We evaluate on seven benchmarks: MATH-500~\citep{math} and GSM8K~\citep{gsm8k} for mathematical reasoning; MBPP~\citep{austin2021program} and HumanEval~\citep{chen2021evaluating} for code generation; Countdown and Sudoku for constraint satisfaction; and JSON schema adherence for constrained generation. The full \ours{} framework (\dps{} + \sml{}) is applied to math, code tasks, and JSON generation(\cref{tab:main-math-code,tab:json-results}) as well as for constrained generation; for constraint satisfaction, only \dps{} is applied (\cref{tab:structured-results}). We evaluate at generation lengths 128, 256, and 512 with proportionally scaled diffusion steps to maintain consistency with \citet{d1,wd1}. For Sudoku, we report accuracy per-empty-cell following~\citet{wd1}. \cref{app:reproducibility,app:ablations} contain more details and analyses.

\begin{table}[ht]
\centering
\caption{Accuracy (\%) on math reasoning and code generation tasks (pass@1) by generation length.}
\label{tab:main-math-code}
\small
\setlength{\tabcolsep}{2.8pt}
\begin{tabular}{@{}l p{4pt} ccc p{4pt} ccc p{4pt} ccc p{4pt} ccc@{}}
 & & \multicolumn{3}{c}{\textbf{MATH-500}} & & \multicolumn{3}{c}{\textbf{GSM8K}} & & \multicolumn{3}{c}{\textbf{MBPP}} & & \multicolumn{3}{c}{\textbf{HumanEval}} \\
\cmidrule(lr){3-5} \cmidrule(lr){7-9} \cmidrule(lr){11-13} \cmidrule(lr){15-17}
\textbf{Method} & & \textbf{128} & \textbf{256} & \textbf{512} & & \textbf{128} & \textbf{256} & \textbf{512} & & \textbf{128} & \textbf{256} & \textbf{512} & & \textbf{128} & \textbf{256} & \textbf{512} \\
\midrule
LLaDA-8B-Inst. & & 26.0 & 32.4 & 36.2 & & 68.7 & 76.7 & 78.2 & & 35.8 & 42.6 & 45.2 & & 17.1 & 26.2 & 34.2 \\
\addlinespace[6pt]
Diffu-GRPO & & 31.8 & 37.1 & 39.2 & & 72.1 & 79.2 & 79.8 & & 41.7 & 44.9 & 44.2 & & 19.8 & 32.3 & 35.4 \\
\rowcolor{ourscolor}
\quad + \dps{} & & 33.5 & 37.5 & 40.2 & & 74.0 & 80.8 & 80.9 & & \textbf{47.0} & 46.5 & 45.6 & & 22.9 & 32.8 & 36.5 \\
\rowcolor{ourscolor}
\quad + \sml{} & & 34.3 & 38.3 & \textbf{40.5} & & 74.6 & \textbf{81.6} & \textbf{81.5} & & 44.4 & 47.5 & 48.6 & & \textbf{26.6} & \textbf{36.5} & \textbf{40.6} \\
\rowcolor{ourscolor}
\quad + \ours{} & & \textbf{35.4} & \textbf{38.8} & \textbf{40.5} & & \textbf{74.6} & \textbf{81.6} & \textbf{81.5} & & 45.4 & \textbf{48.1} & \textbf{50.5} & & 25.5 & 35.4 & 40.1 \\
\addlinespace[6pt]
wd1 & & 32.6 & 36.7 & 39.6 & & 70.5 & 80.6 & 80.7 & & 45.4 & 45.8 & 45.8 & & 21.9 & 34.9 & 37.5 \\
\rowcolor{ourscolor}
\quad + \dps{} & & 33.5 & 39.8 & 40.9 & & 72.8 & 81.6 & 81.6 & & \textbf{46.1} & 47.5 & 47.5 & & 25.0 & \textbf{36.5} & 40.0 \\
\rowcolor{ourscolor}
\quad + \sml{} & & 34.3 & 39.4 & \textbf{41.5} & & 74.9 & 81.9 & 82.0 & & 45.8 & 47.2 & 47.2 & & \textbf{25.5} & 34.9 & 39.1 \\
\rowcolor{ourscolor}
\quad + \ours{} & & \textbf{35.6} & \textbf{40.2} & 41.3 & & \textbf{76.1} & \textbf{82.1} & \textbf{82.1} & & 45.4 & \textbf{48.6} & \textbf{50.7} & & 25.0 & 35.9 & \textbf{40.1} \\
\addlinespace[6pt]
GDPO & & 33.9 & 36.9 & 40.7 & & 73.6 & 80.6 & 80.0 & & 38.9 & 45.1 & 46.1 & & 18.8 & 37.0 & 37.0 \\
\rowcolor{ourscolor}
\quad + \dps{} & & 34.9 & \textbf{39.0} & 40.7 & & \textbf{75.1} & 80.6 & \textbf{82.8} & & 40.7 & 47.7 & 48.2 & & 21.9 & 35.9 & 40.0 \\
\rowcolor{ourscolor}
\quad + \sml{} & & 34.5 & \textbf{39.0} & 40.2 & & 74.0 & 80.6 & 80.6 & & 45.8 & 47.5 & \textbf{50.0} & & \textbf{25.5} & 36.5 & 40.6 \\
\rowcolor{ourscolor}
\quad + \ours{} & & \textbf{35.2} & 38.6 & \textbf{41.7} & & 74.3 & \textbf{81.6} & 82.0 & & \textbf{46.3} & \textbf{48.1} & 49.5 & & 25.0 & \textbf{37.0} & \textbf{41.1} \\
\end{tabular}
\end{table}

\textbf{Main Results.}
\label{sec:main-results}
\cref{tab:main-math-code} presents mathematical reasoning (MATH-500, GSM8K) and code generation results (MBPP, HumanEval). Three patterns emerge.

\textbf{DPS and SML are complementary.} \ours{} achieves the best or tied-best average improvement in 9 of 12 method--benchmark combinations. From \cref{tab:main-math-code} we see that \sml{} and \dps{} are complementary; both methods lead to increases and together (with \ours{}) we see near additive results. For example, on wd1 MBPP, \dps{} alone adds +1.4pp and \sml{} alone adds +1.1pp, but \ours{} adds +2.6pp---more than either component and close to their sum. This confirms \ours{} is usually the best choice.

\textbf{The two mechanisms target different regimes.} \dps{} contributes most at short generation lengths, where each denoising step resolves a larger fraction of the output and the distinction between decisive and routine steps is sharpest (e.g., +5.3pp on MBPP at $L{=}128$ with d1). \sml{} contributes most at long generation lengths, where inter-token dependencies are stronger and the mean-field bias is more severe (e.g., +4.4pp on MBPP at $L{=}512$ with d1; +5.2pp on HumanEval at $L{=}512$ with d1). \ours{} captures both effects, achieving the highest MBPP accuracy at $L{=}512$ across all three base methods (50.5, 50.7, and 49.5 respectively).

\textbf{Improvements are consistent across base methods.} Each individual component improves nearly every method--benchmark combination regardless of whether the base method is ratio-based (d1, GDPO) or ratio-free (wd1), and regardless of whether it operates at the token level (d1, wd1) or sequence level (GDPO). This validates the plug-and-play design: \dps{} and \sml{} address upstream weaknesses in the training signal that are shared across loss formulations.

\textbf{\sml{} Applicability: Task Structure and Output Length}.
\label{sec:sml-limitations}
We apply \sml{} only to long-form tasks (MATH-500, GSM8K, MBPP, HumanEval) and excluded from constraint satisfaction (Countdown, Sudoku) due to a distributional mismatch. \sml{} evaluates each token with $(K{-}1)/K$ visible context (75\% for $K{=}4$), whereas generation starts fully masked. For short, tightly constrained outputs (${\sim}7$ empty Sudoku cells, 10--15 Countdown tokens), 75\% context nearly determines the answer via constraint propagation, driving the model toward trivial gap-filling that fails at generation.

\begin{wraptable}{r}{0.52\textwidth}
	\vspace{-14pt}
	\centering
	\caption{\textbf{Constraint satisfaction (\dps{} only).} Accuracy (\%) on Countdown and Sudoku. Sudoku uses per-cell accuracy~\citep{wd1}.}
	\label{tab:structured-results}
	\small
	\setlength{\tabcolsep}{3pt}
	\begin{tabular}{@{}lccccccc@{}}
		& \multicolumn{3}{c}{\textbf{Countdown}} & & \multicolumn{3}{c}{\textbf{Sudoku}} \\
		\cmidrule(lr){2-4} \cmidrule(lr){6-8}
		\textbf{Method} & \textbf{128} & \textbf{256} & \textbf{512} & & \textbf{128} & \textbf{256} & \textbf{512} \\
		\midrule
		LLaDA-8B-Inst. & 20.7 & 19.5 & 16.0 & & 11.7 & 6.7 & 5.5 \\
		\addlinespace[6pt]
		Diffu-GRPO & 50.0 & 50.3 & 47.9 & & 25.9 & 26.2 & 26.3 \\
		\rowcolor{ourscolor}
		\quad + \dps{} & \textbf{57.6} & \textbf{55.2} & \textbf{53.8} & & \textbf{26.8} & \textbf{29.2} & \textbf{37.5} \\
		\addlinespace[6pt]
		wd1 & 53.1 & 53.1 & 47.9 & & 51.4 & 40.5 & 34.6 \\
		\rowcolor{ourscolor}
		\quad + \dps{} & \textbf{54.3} & \textbf{56.9} & \textbf{58.3} & & \textbf{77.6} & \textbf{76.1} & \textbf{70.9} \\
		\addlinespace[6pt]
		GDPO & \textbf{62.5} & 69.4 & 74.7 & & 26.1 & 26.2 & 25.8 \\
		\rowcolor{ourscolor}
		\quad + \dps{} & 60.1 & \textbf{70.2} & \textbf{76.8} & & \textbf{38.9} & \textbf{50.0} & \textbf{37.5} \\
	\end{tabular}
	\vspace{-8pt}
\end{wraptable}

\paragraph{Constraint satisfaction analysis.} \dps{} delivers the most dramatic improvements on Sudoku, where wd1 + \dps{} improves per-cell accuracy by +26--36pp (\cref{tab:structured-results})---consistent with the task's tight inter-cell dependencies, which make decisive denoising steps especially impactful. The jump reflects a qualitative shift: wd1's per-cell accuracy at $L{=}512$ rises from 34.6\% to 70.9\%, crossing from unreliable to reliable. On Countdown, \dps{} improves d1 and wd1 consistently (+1.2--10.4pp) but shows a small degradation on GDPO at $L{=}128$ ($-2.4$pp), likely because GDPO's stronger baseline leaves less headroom.

\paragraph{JSON generation.} \dps{} improves schema adherence for all three base methods (\Cref{fig:dps-results}). The task is non-trivial even after RL post-training: the strongest base method (wd1 at $L{=}512$) reaches only 72.57\% schema adherence on \textsc{github-medium}, reflecting the difficulty of generating nested structures that fully conform to a target schema. Combining both components amplifies the effect: \ours{} achieves the highest schema adherence on wd1 (+5.90pp, reaching 78.47\%) and GDPO (+3.82pp), with \dps{} and \sml{} contributing comparably on wd1 alone (+3.47pp and +3.82pp). d1 is the lone exception: \dps{} alone (+4.52pp) narrowly beats \ours{} (+4.17pp), suggesting mild interference between the two components. Despite JSON's rigid schema structure, \sml{} is effective here---unlike on Sudoku and Countdown---because outputs at $L{=}512$ are long enough that partial-mask context does not trivially determine the remaining tokens. This sharpens \sml{}'s applicability boundary as length-driven rather than structure-driven; see \cref{app:json-generation} for the full table and analysis. 
\section{Discussion and conclusion}
\label{sec:conclusion}

We presented \ours{}, a plug-and-play framework that brings temporal credit assignment and improved likelihood estimation to any GRPO-style dLLM trainer, with consistent improvement across seven benchmarks and three base methods.

The most striking result is that ``wasted'' intermediate predictions---logits that existing methods compute and discard at every denoising step---contain enough signal to meaningfully improve training. \dps{} exploits this signal at near-zero cost, and the dramatic Sudoku gains (+26--36pp) suggest that structured tasks with tight inter-token dependencies are especially under-served by uniform credit assignment. More broadly, the fact that a simple per-step importance measure transfers across ratio-based and ratio-free losses, across math, code, and constraint satisfaction domains, and across short and long generation lengths indicates that the missing credit-assignment structure is a shared bottleneck in current dLLM RL pipelines---not an artifact of any single loss formulation.

\sml{} complements \dps{} for long-form tasks where mean-field bias is most severe, but its value diminishes when the base method already diversifies likelihood views (as with GDPO's SDMC). This points to a practical guideline: \dps{} is a universal add-on, while \sml{} should be prioritized for methods and tasks where likelihood quality is the binding constraint.

\textbf{Limitations.} \sml{} is not applicable to short, tightly constrained outputs, where the distributional mismatch between partial-mask training and fully-masked generation causes mode collapse; developing a likelihood regularizer effective across all output lengths remains open. 

\textbf{Future directions.} Adaptive stride selection---adjusting recording frequency based on observed delta magnitudes---could improve the signal-to-noise tradeoff without manual tuning. Finally, \sml{}'s stratified estimation principle may extend beyond RL to the pre-training objective itself, where the same mean-field bias affects the masked diffusion loss.

\bibliographystyle{plainnat}
\bibliography{references}

\begin{thebibliography}{30}
\providecommand{\natexlab}[1]{#1}
\providecommand{\url}[1]{\texttt{#1}}
\expandafter\ifx\csname urlstyle\endcsname\relax
  \providecommand{\doi}[1]{doi: #1}\else
  \providecommand{\doi}{doi: \begingroup \urlstyle{rm}\Url}\fi

\bibitem[Austin et~al.(2021)Austin, Odena, Nye, Bosma, Michalewski, Dohan, Jiang, Cai, Terry, Le, and Sutton]{austin2021program}
Jacob Austin, Augustus Odena, Maxwell Nye, Maarten Bosma, Henryk Michalewski, David Dohan, Ellen Jiang, Carrie Cai, Michael Terry, Quoc Le, and Charles Sutton.
\newblock Program synthesis with large language models, 2021.
\newblock URL \url{https://arxiv.org/abs/2108.07732}.

\bibitem[Black et~al.(2024)Black, Janner, Du, Kostrikov, and Levine]{ddpo}
Kevin Black, Michael Janner, Yilun Du, Ilya Kostrikov, and Sergey Levine.
\newblock Training diffusion models with reinforcement learning.
\newblock In \emph{The Twelfth International Conference on Learning Representations}, 2024.
\newblock URL \url{https://openreview.net/forum?id=YCWjhGrJFD}.

\bibitem[Chen et~al.(2021)Chen, Tworek, Jun, Yuan, de~Oliveira~Pinto, Kaplan, Edwards, Burda, Joseph, Brockman, Ray, Puri, Krueger, Petrov, Khlaaf, Sastry, Mishkin, Chan, Gray, Ryder, Pavlov, Power, Kaiser, Bavarian, Winter, Tillet, Such, Cummings, Plappert, Chantzis, Barnes, Herbert-Voss, Guss, Nichol, Paino, Tezak, Tang, Babuschkin, Balaji, Jain, Saunders, Hesse, Carr, Leike, Achiam, Misra, Morikawa, Radford, Knight, Brundage, Murati, Mayer, Welinder, McGrew, Amodei, McCandlish, Sutskever, and Zaremba]{chen2021evaluating}
Mark Chen, Jerry Tworek, Heewoo Jun, Qiming Yuan, Henrique~Ponde de~Oliveira~Pinto, Jared Kaplan, Harri Edwards, Yuri Burda, Nicholas Joseph, Greg Brockman, Alex Ray, Raul Puri, Gretchen Krueger, Michael Petrov, Heidy Khlaaf, Girish Sastry, Pamela Mishkin, Brooke Chan, Scott Gray, Nick Ryder, Mikhail Pavlov, Alethea Power, Lukasz Kaiser, Mohammad Bavarian, Clemens Winter, Philippe Tillet, Felipe~Petroski Such, Dave Cummings, Matthias Plappert, Fotios Chantzis, Elizabeth Barnes, Ariel Herbert-Voss, William~Hebgen Guss, Alex Nichol, Alex Paino, Nikolas Tezak, Jie Tang, Igor Babuschkin, Suchir Balaji, Shantanu Jain, William Saunders, Christopher Hesse, Andrew~N. Carr, Jan Leike, Josh Achiam, Vedant Misra, Evan Morikawa, Alec Radford, Matthew Knight, Miles Brundage, Mira Murati, Katie Mayer, Peter Welinder, Bob McGrew, Dario Amodei, Sam McCandlish, Ilya Sutskever, and Wojciech Zaremba.
\newblock Evaluating large language models trained on code, 2021.
\newblock URL \url{https://arxiv.org/abs/2107.03374}.

\bibitem[Cobbe et~al.(2021)Cobbe, Kosaraju, Bavarian, Chen, Jun, Kaiser, Plappert, Tworek, Hilton, Nakano, Hesse, and Schulman]{gsm8k}
Karl Cobbe, Vineet Kosaraju, Mohammad Bavarian, Mark Chen, Heewoo Jun, Lukasz Kaiser, Matthias Plappert, Jerry Tworek, Jacob Hilton, Reiichiro Nakano, Christopher Hesse, and John Schulman.
\newblock Training verifiers to solve math word problems, 2021.
\newblock URL \url{https://arxiv.org/abs/2110.14168}.

\bibitem[Feng et~al.(2026)Feng, Ma, Nan, Chen, Zhai, Griffiths, Gao, Gan, Verma, Yang, Chen, and Dehghan]{so-bench}
Di~Feng, Kaixin Ma, Feng Nan, Haofeng Chen, Bohan Zhai, David Griffiths, Mingfei Gao, Zhe Gan, Eshan Verma, Yinfei Yang, Zhifeng Chen, and Afshin Dehghan.
\newblock So-bench: A structural output evaluation of multimodal llms, 2026.
\newblock URL \url{https://arxiv.org/abs/2511.21750}.

\bibitem[Geng et~al.(2025)Geng, Cooper, Moskal, Jenkins, Berman, Ranchin, West, Horvitz, and Nori]{JSONSchemaBench}
Saibo Geng, Hudson Cooper, Micha{\l} Moskal, Samuel Jenkins, Julian Berman, Nathan Ranchin, Robert West, Eric Horvitz, and Harsha Nori.
\newblock Jsonschemabench: A rigorous benchmark of structured outputs for language models, 2025.
\newblock URL \url{https://arxiv.org/abs/2501.10868}.

\bibitem[Gong et~al.(2025)Gong, Zhang, Zheng, Gu, Jaitly, Kong, and Zhang]{gong2025diffucoder}
Shansan Gong, Ruixiang Zhang, Huangjie Zheng, Jiatao Gu, Navdeep Jaitly, Lingpeng Kong, and Yizhe Zhang.
\newblock Diffucoder: Understanding and improving masked diffusion models for code generation.
\newblock \emph{arXiv preprint arXiv:2506.20639}, 2025.
\newblock URL \url{https://arxiv.org/abs/2506.20639}.

\bibitem[Guo et~al.(2025)Guo, Yang, Zhang, Song, Wang, Zhu, Xu, Zhang, Ma, Bi, Zhang, Yu, Wu, Wu, Gou, Shao, Li, Gao, Liu, Xue, Wang, Wu, Feng, Lu, Zhao, Deng, Ruan, Dai, Chen, Ji, Li, Lin, Dai, Luo, Hao, Chen, Li, Zhang, Xu, Ding, Gao, Qu, Li, Guo, Li, Chen, Yuan, Tu, Qiu, Li, Cai, Ni, Liang, Chen, Dong, Hu, You, Gao, Guan, Huang, Yu, Wang, Zhang, Zhao, Wang, Zhang, Xu, Xia, Zhang, Zhang, Tang, Zhou, Li, Wang, Li, Tian, Huang, Zhang, Wang, Chen, Du, Ge, Zhang, Pan, Wang, Chen, Jin, Chen, Lu, Zhou, Chen, Ye, Wang, Yu, Zhou, Pan, Li, Zhou, Wu, Yun, Pei, Sun, Wang, Zeng, Liu, Liang, Gao, Yu, Zhang, Xiao, An, Liu, Wang, Chen, Nie, Cheng, Liu, Xie, Liu, Yang, Li, Su, Lin, Li, Jin, Shen, Chen, Sun, Wang, Song, Zhou, Wang, Shan, Li, Wang, Wei, Zhang, Xu, Li, Zhao, Sun, Wang, Yu, Zhang, Shi, Xiong, He, Piao, Wang, Tan, Ma, Liu, Guo, Ou, Wang, Gong, Zou, He, Xiong, Luo, You, Liu, Zhou, Zhu, Huang, Li, Zheng, Zhu, Ma, Tang, Zha, Yan, Ren, Ren, Sha, Fu, Xu, Xie, Zhang, Hao, Ma, Yan, Wu, Gu, Zhu, Liu, Li, Xie, Song,
  Pan, Huang, Xu, Zhang, and Zhang]{guo2025deepseek}
Daya Guo, Dejian Yang, Haowei Zhang, Junxiao Song, Peiyi Wang, Qihao Zhu, Runxin Xu, Ruoyu Zhang, Shirong Ma, Xiao Bi, Xiaokang Zhang, Xingkai Yu, Yu~Wu, Z.~F. Wu, Zhibin Gou, Zhihong Shao, Zhuoshu Li, Ziyi Gao, Aixin Liu, Bing Xue, Bingxuan Wang, Bochao Wu, Bei Feng, Chengda Lu, Chenggang Zhao, Chengqi Deng, Chong Ruan, Damai Dai, Deli Chen, Dongjie Ji, Erhang Li, Fangyun Lin, Fucong Dai, Fuli Luo, Guangbo Hao, Guanting Chen, Guowei Li, H.~Zhang, Hanwei Xu, Honghui Ding, Huazuo Gao, Hui Qu, Hui Li, Jianzhong Guo, Jiashi Li, Jingchang Chen, Jingyang Yuan, Jinhao Tu, Junjie Qiu, Junlong Li, J.~L. Cai, Jiaqi Ni, Jian Liang, Jin Chen, Kai Dong, Kai Hu, Kaichao You, Kaige Gao, Kang Guan, Kexin Huang, Kuai Yu, Lean Wang, Lecong Zhang, Liang Zhao, Litong Wang, Liyue Zhang, Lei Xu, Leyi Xia, Mingchuan Zhang, Minghua Zhang, Minghui Tang, Mingxu Zhou, Meng Li, Miaojun Wang, Mingming Li, Ning Tian, Panpan Huang, Peng Zhang, Qiancheng Wang, Qinyu Chen, Qiushi Du, Ruiqi Ge, Ruisong Zhang, Ruizhe Pan, Runji Wang, R.~J.
  Chen, R.~L. Jin, Ruyi Chen, Shanghao Lu, Shangyan Zhou, Shanhuang Chen, Shengfeng Ye, Shiyu Wang, Shuiping Yu, Shunfeng Zhou, Shuting Pan, S.~S. Li, Shuang Zhou, Shaoqing Wu, Tao Yun, Tian Pei, Tianyu Sun, T.~Wang, Wangding Zeng, Wen Liu, Wenfeng Liang, Wenjun Gao, Wenqin Yu, Wentao Zhang, W.~L. Xiao, Wei An, Xiaodong Liu, Xiaohan Wang, Xiaokang Chen, Xiaotao Nie, Xin Cheng, Xin Liu, Xin Xie, Xingchao Liu, Xinyu Yang, Xinyuan Li, Xuecheng Su, Xuheng Lin, X.~Q. Li, Xiangyue Jin, Xiaojin Shen, Xiaosha Chen, Xiaowen Sun, Xiaoxiang Wang, Xinnan Song, Xinyi Zhou, Xianzu Wang, Xinxia Shan, Y.~K. Li, Y.~Q. Wang, Y.~X. Wei, Yang Zhang, Yanhong Xu, Yao Li, Yao Zhao, Yaofeng Sun, Yaohui Wang, Yi~Yu, Yichao Zhang, Yifan Shi, Yiliang Xiong, Ying He, Yishi Piao, Yisong Wang, Yixuan Tan, Yiyang Ma, Yiyuan Liu, Yongqiang Guo, Yuan Ou, Yuduan Wang, Yue Gong, Yuheng Zou, Yujia He, Yunfan Xiong, Yuxiang Luo, Yuxiang You, Yuxuan Liu, Yuyang Zhou, Y.~X. Zhu, Yanping Huang, Yaohui Li, Yi~Zheng, Yuchen Zhu, Yunxian Ma, Ying
  Tang, Yukun Zha, Yuting Yan, Z.~Z. Ren, Zehui Ren, Zhangli Sha, Zhe Fu, Zhean Xu, Zhenda Xie, Zhengyan Zhang, Zhewen Hao, Zhicheng Ma, Zhigang Yan, Zhiyu Wu, Zihui Gu, Zijia Zhu, Zijun Liu, Zilin Li, Ziwei Xie, Ziyang Song, Zizheng Pan, Zhen Huang, Zhipeng Xu, Zhongyu Zhang, and Zhen Zhang.
\newblock Deepseek-r1 incentivizes reasoning in llms through reinforcement learning.
\newblock \emph{Nature}, 645\penalty0 (8081):\penalty0 633–638, September 2025.
\newblock ISSN 1476-4687.
\newblock \doi{10.1038/s41586-025-09422-z}.
\newblock URL \url{http://dx.doi.org/10.1038/s41586-025-09422-z}.

\bibitem[Hendrycks et~al.(2021)Hendrycks, Burns, Kadavath, Arora, Basart, Tang, Song, and Steinhardt]{math}
Dan Hendrycks, Collin Burns, Saurav Kadavath, Akul Arora, Steven Basart, Eric Tang, Dawn Song, and Jacob Steinhardt.
\newblock Measuring mathematical problem solving with the {MATH} dataset.
\newblock In \emph{Thirty-fifth Conference on Neural Information Processing Systems Datasets and Benchmarks Track (Round 2)}, 2021.
\newblock URL \url{https://openreview.net/forum?id=7Bywt2mQsCe}.

\bibitem[Labs et~al.(2025)Labs, Khanna, Kharbanda, Li, Varma, Wang, Birnbaum, Luo, Miraoui, Palrecha, Ermon, Grover, and Kuleshov]{khanna2025mercury}
Inception Labs, Samar Khanna, Siddhant Kharbanda, Shufan Li, Harshit Varma, Eric Wang, Sawyer Birnbaum, Ziyang Luo, Yanis Miraoui, Akash Palrecha, Stefano Ermon, Aditya Grover, and Volodymyr Kuleshov.
\newblock Mercury: Ultra-fast language models based on diffusion, 2025.
\newblock URL \url{https://arxiv.org/abs/2506.17298}.

\bibitem[Lee et~al.(2026)Lee, Lee, Dan, Park, and Ahn]{leeDyLLMEfficientDiffusion2026}
Younjoo Lee, Junghoo Lee, Seungkyun Dan, Jaiyoung Park, and Jung~Ho Ahn.
\newblock Dyllm: Efficient diffusion llm inference via saliency-based token selection and partial attention, 2026.
\newblock URL \url{https://arxiv.org/abs/2603.08026}.

\bibitem[Lightman et~al.(2024)Lightman, Kosaraju, Burda, Edwards, Baker, Lee, Leike, Schulman, Sutskever, and Cobbe]{prm}
Hunter Lightman, Vineet Kosaraju, Yuri Burda, Harrison Edwards, Bowen Baker, Teddy Lee, Jan Leike, John Schulman, Ilya Sutskever, and Karl Cobbe.
\newblock Let's verify step by step.
\newblock In \emph{The Twelfth International Conference on Learning Representations}, 2024.
\newblock URL \url{https://openreview.net/forum?id=v8L0pN6EOi}.

\bibitem[Liu et~al.(2025)Liu, Chen, Li, Qi, Pang, Du, Lee, and Lin]{liu2025understanding}
Zichen Liu, Changyu Chen, Wenjun Li, Penghui Qi, Tianyu Pang, Chao Du, Wee~Sun Lee, and Min Lin.
\newblock Understanding r1-zero-like training: A critical perspective, 2025.
\newblock URL \url{https://arxiv.org/abs/2503.20783}.

\bibitem[Lou et~al.(2024)Lou, Meng, and Ermon]{lou2023discrete}
Aaron Lou, Chenlin Meng, and Stefano Ermon.
\newblock Discrete diffusion modeling by estimating the ratios of the data distribution.
\newblock In \emph{Proceedings of the 41st International Conference on Machine Learning}, ICML'24. JMLR.org, 2024.

\bibitem[Monsefi et~al.(2026)Monsefi, Bhendawade, Ciosici, Culver, Zhang, and Belousova]{monsefi2025fs}
Amin~Karimi Monsefi, Nikhil Bhendawade, Manuel~Rafael Ciosici, Dominic Culver, Yizhe Zhang, and Irina Belousova.
\newblock {FS}-{DFM}: Fast and accurate long text generation with few-step diffusion language models.
\newblock In \emph{The Fourteenth International Conference on Learning Representations}, 2026.
\newblock URL \url{https://openreview.net/forum?id=ue1zFeD275}.

\bibitem[M{\"u}ndler et~al.(2026)M{\"u}ndler, Dekoninck, and Vechev]{mundler2026constrained}
Niels M{\"u}ndler, Jasper Dekoninck, and Martin Vechev.
\newblock Constrained decoding of diffusion {LLM}s with context-free grammars.
\newblock In \emph{The Fourteenth International Conference on Learning Representations}, 2026.
\newblock URL \url{https://openreview.net/forum?id=7Sph4KyeYO}.

\bibitem[Nie et~al.(2026)Nie, Zhu, You, Zhang, Ou, Hu, Zhou, Lin, Wen, and Li]{llada}
Shen Nie, Fengqi Zhu, Zebin You, Xiaolu Zhang, Jingyang Ou, Jun Hu, Jun Zhou, Yankai Lin, Ji-Rong Wen, and Chongxuan Li.
\newblock Large language diffusion models.
\newblock In \emph{The Thirty-ninth Annual Conference on Neural Information Processing Systems}, 2026.
\newblock URL \url{https://openreview.net/forum?id=KnqiC0znVF}.

\bibitem[Ouyang et~al.(2022)Ouyang, Wu, Jiang, Almeida, Wainwright, Mishkin, Zhang, Agarwal, Slama, Ray, Schulman, Hilton, Kelton, Miller, Simens, Askell, Welinder, Christiano, Leike, and Lowe]{rlhf}
Long Ouyang, Jeffrey Wu, Xu~Jiang, Diogo Almeida, Carroll Wainwright, Pamela Mishkin, Chong Zhang, Sandhini Agarwal, Katarina Slama, Alex Ray, John Schulman, Jacob Hilton, Fraser Kelton, Luke Miller, Maddie Simens, Amanda Askell, Peter Welinder, Paul~F Christiano, Jan Leike, and Ryan Lowe.
\newblock Training language models to follow instructions with human feedback.
\newblock In S.~Koyejo, S.~Mohamed, A.~Agarwal, D.~Belgrave, K.~Cho, and A.~Oh, editors, \emph{Advances in Neural Information Processing Systems}, volume~35, pages 27730--27744. Curran Associates, Inc., 2022.
\newblock URL \url{https://proceedings.neurips.cc/paper_files/paper/2022/file/b1efde53be364a73914f58805a001731-Paper-Conference.pdf}.

\bibitem[Rojas et~al.(2026)Rojas, Lin, Rasul, Schneider, Nevmyvaka, Tao, and Deng]{rojas2025improving}
Kevin Rojas, Jiahe Lin, Kashif Rasul, Anderson Schneider, Yuriy Nevmyvaka, Molei Tao, and Wei Deng.
\newblock Improving reasoning for diffusion language models via group diffusion policy optimization, 2026.
\newblock URL \url{https://arxiv.org/abs/2510.08554}.

\bibitem[Sahoo et~al.(2024)Sahoo, Arriola, Schiff, Gokaslan, Marroquin, Chiu, Rush, and Kuleshov]{mdlm}
Subham~Sekhar Sahoo, Marianne Arriola, Yair Schiff, Aaron Gokaslan, Edgar Marroquin, Justin~T Chiu, Alexander Rush, and Volodymyr Kuleshov.
\newblock Simple and effective masked diffusion language models.
\newblock In A.~Globerson, L.~Mackey, D.~Belgrave, A.~Fan, U.~Paquet, J.~Tomczak, and C.~Zhang, editors, \emph{Advances in Neural Information Processing Systems}, volume~37, pages 130136--130184. Curran Associates, Inc., 2024.
\newblock \doi{10.52202/079017-4135}.
\newblock URL \url{https://proceedings.neurips.cc/paper_files/paper/2024/file/eb0b13cc515724ab8015bc978fdde0ad-Paper-Conference.pdf}.

\bibitem[Shao et~al.(2024)Shao, Wang, Zhu, Xu, Song, Bi, Zhang, Zhang, Li, Wu, and Guo]{grpo}
Zhihong Shao, Peiyi Wang, Qihao Zhu, Runxin Xu, Junxiao Song, Xiao Bi, Haowei Zhang, Mingchuan Zhang, Y.~K. Li, Y.~Wu, and Daya Guo.
\newblock Deepseekmath: Pushing the limits of mathematical reasoning in open language models, 2024.
\newblock URL \url{https://arxiv.org/abs/2402.03300}.

\bibitem[Sutton and Barto(2018)]{sutton_barto}
Richard~S Sutton and Andrew~G Barto.
\newblock \emph{Reinforcement Learning: An Introduction}.
\newblock MIT press, 2018.

\bibitem[Tang et~al.(2026)Tang, Dolga, Yoon, and Bogunovic]{wd1}
Xiaohang Tang, Rares Dolga, Sangwoong Yoon, and Ilija Bogunovic.
\newblock wd1: Weighted policy optimization for reasoning in diffusion language models.
\newblock \emph{ICLR}, 2026.
\newblock URL \url{https://openreview.net/forum?id=L2rfd2Czbj}.

\bibitem[Uesato et~al.(2022)Uesato, Kushman, Kumar, Song, Siegel, Wang, Creswell, Irving, and Higgins]{process_reward}
Jonathan Uesato, Nate Kushman, Ramana Kumar, Francis Song, Noah Siegel, Lisa Wang, Antonia Creswell, Geoffrey Irving, and Irina Higgins.
\newblock Solving math word problems with process- and outcome-based feedback, 2022.
\newblock URL \url{https://arxiv.org/abs/2211.14275}.

\bibitem[Xie et~al.(2025)Xie, Ye, Zheng, Gao, Dong, Wu, Zhao, Gong, Jiang, Li, and Kong]{xie2025dream}
Zhihui Xie, Jiacheng Ye, Lin Zheng, Jiahui Gao, Jingwei Dong, Zirui Wu, Xueliang Zhao, Shansan Gong, Xin Jiang, Zhenguo Li, and Lingpeng Kong.
\newblock Dream-coder 7b: An open diffusion language model for code, 2025.
\newblock URL \url{https://arxiv.org/abs/2509.01142}.

\bibitem[Xu et~al.(2025)Xu, Liu, Yin, Zhou, and Poovendran]{kodcode}
Zhangchen Xu, Yang Liu, Yueqin Yin, Mingyuan Zhou, and Radha Poovendran.
\newblock Kodcode: A diverse, challenging, and verifiable synthetic dataset for coding, 2025.
\newblock URL \url{https://arxiv.org/abs/2503.02951}.

\bibitem[Ye et~al.(2025)Ye, Xie, Zheng, Gao, Wu, Jiang, Li, and Kong]{ye2025dream}
Jiacheng Ye, Zhihui Xie, Lin Zheng, Jiahui Gao, Zirui Wu, Xin Jiang, Zhenguo Li, and Lingpeng Kong.
\newblock Dream 7b: Diffusion large language models, 2025.
\newblock URL \url{https://arxiv.org/abs/2508.15487}.

\bibitem[Yu et~al.(2025)Yu, Zhang, Zhu, Yuan, Zuo, Yue, Dai, Fan, Liu, Liu, Liu, Lin, Lin, Ma, Sheng, Tong, Zhang, Zhang, Zhang, Zhu, Zhu, Chen, Chen, Wang, Yu, Song, Wei, Zhou, Liu, Ma, Zhang, Yan, Qiao, Wu, and Wang]{yu2025dapo}
Qiying Yu, Zheng Zhang, Ruofei Zhu, Yufeng Yuan, Xiaochen Zuo, Yu~Yue, Weinan Dai, Tiantian Fan, Gaohong Liu, Lingjun Liu, Xin Liu, Haibin Lin, Zhiqi Lin, Bole Ma, Guangming Sheng, Yuxuan Tong, Chi Zhang, Mofan Zhang, Wang Zhang, Hang Zhu, Jinhua Zhu, Jiaze Chen, Jiangjie Chen, Chengyi Wang, Hongli Yu, Yuxuan Song, Xiangpeng Wei, Hao Zhou, Jingjing Liu, Wei-Ying Ma, Ya-Qin Zhang, Lin Yan, Mu~Qiao, Yonghui Wu, and Mingxuan Wang.
\newblock Dapo: An open-source llm reinforcement learning system at scale, 2025.
\newblock URL \url{https://arxiv.org/abs/2503.14476}.

\bibitem[Zhao et~al.(2025)Zhao, Gupta, Zheng, and Grover]{d1}
Siyan Zhao, Devaansh Gupta, Qinqing Zheng, and Aditya Grover.
\newblock d1: Scaling reasoning in diffusion large language models via reinforcement learning.
\newblock \emph{ICLR}, 2025.
\newblock URL \url{https://openreview.net/forum?id=t8oYNHAvM9}.

\bibitem[Zhu et~al.(2025)Zhu, Xia, Wei, Chen, Chen, and Meng]{zhu2025surprising}
Xinyu Zhu, Mengzhou Xia, Zhepei Wei, Wei-Lin Chen, Danqi Chen, and Yu~Meng.
\newblock The surprising effectiveness of negative reinforcement in llm reasoning.
\newblock \emph{arXiv preprint arXiv:2506.01347}, 2025.

\end{thebibliography}

\crefalias{section}{appendix}
\crefalias{subsection}{appendix}

\appendix

\clearpage
\appendix

\part*{Appendix}
\addcontentsline{toc}{part}{Appendix}

\etocsettocstyle{\section*{Appendix Contents}}{}
\etocsetnexttocdepth{subsection}
\localtableofcontents

\newpage

\section{Experimental details and reproducibility}
\label{app:reproducibility}

\subsection{Hyperparameters}
\label{app:hyperparams}

\cref{tab:hyperparams-training} summarizes the training hyperparameters shared across all experiments. These settings are held constant across all three base methods (Diffu-GRPO, wd1, GDPO) and our extensions, ensuring that any performance differences are attributable to the method rather than training configuration. \cref{tab:hyperparams-dnc} lists the hyperparameters specific to \dps{} and \sml{}.

\begin{table}[]
\centering
\caption{\textbf{Training hyperparameters.} Settings shared across all base methods (Diffu-GRPO, wd1, GDPO) and our extensions.}
\label{tab:hyperparams-training}
\small
\begin{tabular}{@{}llc@{}}
\textbf{Category} & \textbf{Hyperparameter} & \textbf{Value} \\
\midrule
\multicolumn{3}{l}{\textit{Model}} \\
& Base model & LLaDA-8B-Instruct \\
& Precision & bfloat16 \\
& Attention implementation & FlashAttention-2 \\
& Quantization & 4-bit (QLoRA) \\
\midrule
\multicolumn{3}{l}{\textit{LoRA}} \\
& Rank ($r$) & 128 \\
& Alpha ($\alpha$) & 64 \\
& Dropout & 0.05 \\
& Task type & CAUSAL\_LM \\
\midrule
\multicolumn{3}{l}{\textit{Generation}} \\
& Group size ($G$) & 6 \\
& Max completion length & 256 tokens \\
& Max prompt length & 200 tokens \\
& Block length & 32 \\
& Diffusion steps & 128 \\
& Remasking strategy & Low confidence \\
& Random masking & True \\
& Generation batch size & 6 \\
\midrule
\multicolumn{3}{l}{\textit{Optimization}} \\
& Optimizer & AdamW \\
& Learning rate & 3e-6 \\
& LR scheduler & Constant with warmup \\
& Warmup ratio & 0.0001 \\
& Adam $\beta_1$ / $\beta_2$ & 0.9 / 0.99 \\
& Weight decay & 0.1 \\
& Max gradient norm & 0.2 \\
& Per-device batch size & 6 \\
& Gradient accumulation steps & 2 \\
& Number of GPUs & 8 \\
& Inner iterations & 12 \\
& Gradient checkpointing & Off \\
\midrule
\multicolumn{3}{l}{\textit{GRPO-specific}} \\
& PPO clip range ($\epsilon$) & 0.5 \\
& KL coefficient ($\beta$) & 0.0 \\
& Prompt masking probability & 0.15 \\
& Sync reference model & False \\
\midrule
\multicolumn{3}{l}{\textit{Evaluation}} \\
& Generation lengths & 128, 256, 512 \\
& Diffusion steps at eval & 64, 128, 256 \\
& Eval frequency & Every 100 steps \\
& Eval batch size (per device) & 6 \\
\midrule
\multicolumn{3}{l}{\textit{Reproducibility}} \\
& Random seed & 42 \\
\end{tabular}
\end{table}

\begin{table}[ht]
\centering
\caption{\textbf{\ours{} hyperparameters.} Settings specific to our proposed \dps{} and \sml{} components. Default values are used in all main experiments unless otherwise noted; ranges are explored in the ablation studies of \cref{app:ablations}.}
\label{tab:hyperparams-dnc}
\small
\begin{tabular}{@{}llcc@{}}
\textbf{Component} & \textbf{Hyperparameter} & \textbf{Default} & \textbf{Range explored} \\
\midrule
\multicolumn{4}{l}{\textit{\dps{} (Denoising Progress Scores)}} \\
& Modulation strength ($\lambda$) & 0.1 & \{0.05, 0.1, 0.2\} \\
& Record stride ($s$) & -- & \{1, 2, 4, 8, 16, 32\} \\
& Normalization mode & Per-step & \{Per-step, Trajectory, Group, None\} \\
& Last-step mode & Extrapolate & \{Raw, Neutral, Mean, Measured, Extrapolate\} \\
\midrule
\multicolumn{4}{l}{\textit{\sml{} (Stratified Masking Likelihood)}} \\
& Number of strata ($K$) & 4 & \{2, 3, 4, 5, 6, 7, 8\} \\
& Regularization weight ($\eta$) & 0.1 & \{0.05, 0.1, 0.2\} \\
& Stratification strategy & Random & \{Random, Confidence-based\} \\
& Gradient checkpointing & Function-level & --- \\
\end{tabular}
\end{table}

\subsection{Benchmark details}
\label{app:benchmarks}

\cref{tab:benchmark-details} summarizes the benchmarks, training configuration, and which \ours{} components are applied. ``Steps'' refers to the number of optimizer updates (each step processes one batch across all GPUs with gradient accumulation). For mathematical reasoning and constraint satisfaction tasks, we follow the dataset splits and training protocols of~\citet{wd1}. For code generation tasks, we follow~\citet{rojas2025improving}; HumanEval and MBPP is evaluation-only and we train on KodCode ~\citep{kodcode}. For JSON generation, we construct a synthetic dataset from JSONSchemaBench~\citep{JSONSchemaBench}.

\begin{table}[ht]
\centering
\caption{\textbf{Training configuration.} Steps denotes optimizer updates. Reward functions follow~\citet{wd1} for math and constraint tasks. \sml{} is excluded from Countdown and Sudoku due to the distributional mismatch discussed in \cref{sec:sml-limitations}.}
\label{tab:benchmark-details}
\small
\begin{tabular}{@{}llcrcc@{}}
\textbf{Training set} & \textbf{Task type} & \textbf{Reward type} & \textbf{Steps} & \textbf{\dps{}} & \textbf{\sml{}} \\
\midrule
MATH-500 & Math reasoning & Format + correctness & 6,000 & \checkmark & \checkmark \\
GSM8K & Math reasoning & Format + correctness & 7,500 & \checkmark & \checkmark \\
KodCode & Code generation & Format + unit tests & 6,000 & \checkmark & \checkmark \\
Countdown & Constraint sat. & Three-level & 7,500 & \checkmark & \ding{55} \\
Sudoku & Constraint sat. & Per-cell fraction & 12,500 & \checkmark & \ding{55} \\
JSONSchemaBench (our synthetic data) & Constrained gen. & Multi-component & 2,500 & \checkmark & \checkmark \\
\end{tabular}
\end{table}

\paragraph{Reward functions.} Following~\citet{wd1}, we use multi-component reward functions for mathematical reasoning tasks. For MATH-500, the reward combines a format component (+1.0 for answer wrapped in \texttt{\textbackslash boxed\{\}} with reasoning, scaled down for partial formatting) and a correctness component (+2.0 if the boxed answer matches the ground truth), yielding scores in $[0.25, 3.0]$. For GSM8K, the reward combines XML structure compliance (+0.125 per correct tag), soft and strict format matching (+0.5 each), integer answer validation (+0.5), and correctness (+2.0 for exact match), yielding scores in $[0, 3.625]$.
For constraint satisfaction tasks, Countdown uses a three-level reward: +1.0 if the arithmetic expression reaches the target using exactly the given numbers, +0.1 if the correct numbers are used but the target is missed, and 0 otherwise. Sudoku uses the fraction of correctly filled empty cells as the reward, focusing on solving accuracy rather than copying given values.
For code generation tasks (MBPP and HumanEval), the reward is multi-component: A binary formatting reward is used (+1.0 if output is wrapped in \texttt{```python} markdown fences), and a continuous reward which is the fraction of passed unit tests. 

\paragraph{JSON generation.} LLaDA-8B-Instruct, without any RL post-training, already achieves 88--96\% schema adherence on existing JSON generation benchmarks such as json-mode-eval and json-mode-eval-extended \cite{mundler2026constrained}. After applying RL, the model achieves nearly 100\%, making these benchmarks unsuitable for evaluating RL for JSON generation. The primary reason for this is that these benchmarks do not have particularly complex schemas (e.g. lacking nested structures). We therefore construct a synthetic dataset from JSONSchemaBench~\citep{JSONSchemaBench}, which provides a curated collection of JSON schemas organized by difficulty (github-easy, github-medium for example). For each schema, we use GPT-5 to generate three distinct scenarios that fit the schema, then for each scenario we prompt GPT-5 again with the scenario and schema to produce the corresponding JSON output. We filter out examples where the generated ground-truth JSON fails to parse or does not conform to the schema, retaining only verified (schema, scenario, JSON) triples. We preserve the original JSONSchemaBench difficulty splits and evaluate on the github-medium difficulty test split\footnote{We intend to publish this dataset to the public at a later date}.

The JSON generation reward combines three components, inspired by SO-Bench~\citep{so-bench}: (1)~a \emph{format reward} (+1.0 if the output uses \texttt{```json} markdown fences, +0.5 if the output is valid JSON without fences, 0 otherwise); (2)~a \emph{strict schema reward} (binary +1.0 if the fenced JSON is schema-compliant, 0 otherwise); and (3)~a \emph{character-level content reward} using hierarchical field-matching accuracy based on normalized Levenshtein distance. The content reward computes field-matching accuracy between the predicted and ground-truth JSON after recursive normalization (sorting dict keys, aligning list elements), squares the score for sharper gradients, and applies a 0.8$\times$ penalty multiplier when the prediction violates the schema.

\section{Additional method details}
\label{app:method-details}

This section provides extended discussion of design choices summarized briefly in the main text.

\subsection{\dps{}: last-step handling}
\label{app:last-step-details}

Tokens born at the final recorded denoising step $T$ have no subsequent snapshot $S(T{+}1, \cdot)$ to compute a delta against (\cref{fig:last-step-problem}). We evaluate five strategies for handling this boundary case:

\begin{figure}[ht]
\centering
\begin{tikzpicture}[
    every node/.style={font=\footnotesize},
    val/.style={minimum width=1.3cm, minimum height=0.5cm, inner sep=0pt,
                draw=#1!40, fill=#1!8, rounded corners=2pt, font=\footnotesize},
  ]
  \node[font=\footnotesize\bfseries, text=cDark] at (-2.4, 0) {$S(k, \cdot)\!:$};
  \foreach \t/\v/\xp/\lab in {0/-3.2/0/0, 1/-2.8/1.6/1, 2/-2.1/3.2/2, 3/-1.5/4.8/{T{-}1}, 4/-0.9/6.4/T} {
    \node[val=cBlue, font=\scriptsize] (s\t) at (\xp, 0) {$\v$};
    \node[font=\tiny\bfseries, text=cBlue] at (\xp, 0.45) {$\lab$};
  }
  \node[font=\footnotesize\bfseries, text=cDark] at (-2.4, -0.7) {$\delta_k\!:$};
  \foreach \t/\v/\xp in {0/0.4/0, 1/0.7/1.6, 2/0.6/3.2, 3/0.6/4.8} {
    \node[val=cGreen, font=\scriptsize] at (\xp, -0.7) {\v};
  }
  \node[val=cRed, font=\scriptsize\bfseries, text=cRed] at (6.4, -0.7) {?};
  \draw[decorate, decoration={brace, amplitude=3pt, mirror}, cGreen, semithick]
    (-0.7, -1.2) -- (5.5, -1.2)
    node[midway, below=4pt, font=\tiny, text=cGreen] {Known: $\delta_k=S(k{+}1, k{+}1)-S(k, k{+}1)$};
  \draw[decorate, decoration={brace, amplitude=3pt, mirror}, cRed, semithick]
    (5.7, -1.2) -- (7.1, -1.2)
    node[midway, below=4pt, font=\tiny, text=cRed] {No $S(T{+}1, \cdot)$};
\end{tikzpicture}
\caption{\textbf{Last-step boundary problem.} All deltas $\delta_k$ are well-defined except at the final recorded step $T$, where no subsequent snapshot exists to compute the difference against.}
\label{fig:last-step-problem}
\end{figure}
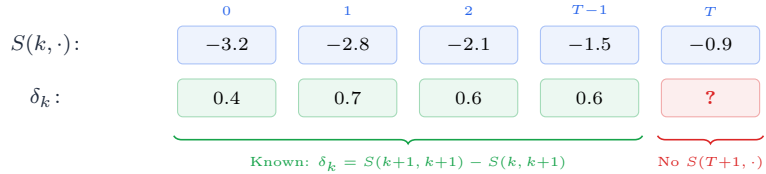

\textbf{Raw Score} sets $\delta_{T} = S(T, T)$, substituting the absolute log-probability score for the missing difference. Because all other deltas are \emph{differences} of scores (\cref{eq:delta}), injecting a raw score introduces a scale mismatch that distorts the per-step normalization in \cref{eq:normalize}.

\textbf{Neutral} sets $\delta_{T} = 0$, effectively discarding the final step's contribution. However, the last denoising steps are where the model resolves its most uncertain tokens; zeroing out their credit wastes precisely the signal that distinguishes difficult tokens from easy ones.

\textbf{Mean} sets $\delta_{T} = \frac{1}{T}\sum_{k<T} \delta_k$, averaging all prior deltas. This is a conservative estimator that smooths step-level noise, but it assumes roughly uniform progress across the denoising trajectory---an assumption at odds with the non-stationary dynamics described in \cref{eq:normalize}.

\textbf{Measured} computes $\delta_{T} = S_{\text{final}} - S(T, T)$, where $S_{\text{final}}$ is obtained by running one additional forward pass on the fully denoised output $\mathbf{x}_0$. This yields the true last-step delta but sacrifices the zero-cost property of \dps{}.

\textbf{Extrapolate} sets $\delta_{T} = \delta_{T-1}$, copying the second-to-last delta under the assumption that adjacent denoising transitions make comparable progress. This preserves scale consistency with other deltas (unlike Raw), retains the last step's credit signal (unlike Neutral), requires no extra computation (unlike Measured), and makes no stationarity assumptions (unlike Mean). As shown in the ablation study (\cref{tab:stride} in \cref{sec:ablation-dps-stride}), Extrapolate is the only mode that achieves the highest average accuracy at both sequence lengths while exceeding the baseline in every stride configuration (6/6). We adopt Extrapolate as the default.

\subsection{\dps{}: recording stride}
\label{app:stride-details}

Computing \dps{} at every denoising step is possible but not necessarily optimal. We record snapshots every $s$ steps (the \emph{record stride}), capturing deltas over coarser intervals. A stride of $s{=}1$ provides the finest granularity but also the noisiest signal: each step unmasks only a few tokens, so the resulting delta $\delta_k$ reflects a small, potentially unrepresentative change in model belief. Larger strides aggregate more token revelations per interval, yielding higher signal-to-noise at the cost of coarser credit assignment---tokens born within the same stride window share a single delta.

The optimal trade-off is inherently task-dependent: tasks with short, structured outputs (e.g., mathematical reasoning) benefit from finer strides that distinguish individual reasoning steps, while tasks with longer, more uniform generations tolerate or even prefer coarser strides where the smoothing effect suppresses spurious fluctuations. We experimented with $s \in \{1, 2, 4, 8, 16, 32\}$ and found that the optimal stride varies across datasets; however, across nearly all configurations, the resulting attribution signal meets or exceeds the baseline---confirming that \dps{} is robust to this hyperparameter. A full ablation over strides and last-step modes is provided in \cref{sec:ablation-dps-stride}.

\subsection{\sml{}: retaining the fully-masked estimate in ratio-based methods}
\label{app:sml-ratio-details}

For ratio-based methods (d1, GDPO), the enriched per-token log-probability (\cref{eq:enriched-ll}) averages each token's fully-masked prediction with its stratified prediction rather than using the stratified estimate alone. Including the fully-masked term is important for two reasons.

First, it prevents a distributional mismatch between training and generation. During generation, the model starts from a fully masked state with zero inter-token context. If the importance ratio is computed entirely from stratified estimates (where each token sees $(K{-}1)/K$ context), the model is optimized under a regime that never occurs during actual inference. The fully-masked term anchors the ratio to the generation-time distribution.

Second, we empirically observe that using only stratified estimates degrades performance. Training becomes unstable at early stages when the model's predictions under partial context diverge significantly from its predictions under no context. The $1/2$ weighting provides a conservative blend that benefits from bias reduction while maintaining alignment with the generation regime.

\subsection{\sml{}: complementarity with prior variance reduction}
\label{app:sml-complementarity}

Several recent methods improve estimation quality in dLLM training, but they target different sources of error at different stages of the pipeline. \sml{} addresses \emph{bias} in the likelihood estimate $\log p_\theta(y)$: given a completed sequence, it controls \emph{which token positions} are masked during evaluation, ensuring each token is predicted with $(K{-}1)/K$ context rather than zero. DiffuCoder's coupled sampling~\citep{gong2025diffucoder} and GDPO's SDMC (\cref{eq:elbo}) instead address \emph{variance} in the diffusion training objective: they control \emph{which timesteps} $t$ are used to estimate the ELBO---via antithetic pairing or deterministic quadrature, respectively. In short, \sml{} operates on the token position axis within a single forward pass, while coupled sampling and SDMC operate on the diffusion time axis across forward passes. The mechanisms do not interfere: \sml{} can serve as a drop-in replacement for the likelihood estimate inside any method that already uses coupled sampling or SDMC.

\section{Ablation studies and sensitivity analysis}
\label{app:ablations}

\ours{} introduces several design choices and hyperparameters across its two components. In this section we systematically evaluate each one, isolating its effect on MATH-500 with wd1 as the base method unless otherwise noted. We organize the analysis into three groups: \dps{} design choices covering the record stride, last-step handling, normalization mode, and modulation strength $\lambda$ (\crefrange{sec:ablation-dps-stride}{app:ablation-lambda}); \sml{} design choices covering the number of strata, stratification strategy, and regularization weight $\eta$ (\crefrange{app:ablation-sml}{app:ablation-eta}); and computational overhead, measuring the GPU memory and wall-clock cost of each component (\cref{app:compute-overhead}).

\subsection{\dps{} record stride and last-step mode}
\label{sec:ablation-dps-stride}

\cref{tab:stride} ablates two coupled design choices: the record stride $s$ and the strategy for computing the final delta $\delta_{T}$, where no subsequent snapshot $S(T{+}1, \cdot)$ exists (see \cref{eq:delta}). We evaluate five last-step modes (see \cref{app:last-step-details} for detailed descriptions):

\begin{table}[t]
\centering
\caption{\textbf{\dps{} record stride and last-step mode ablation} on MATH-500 with wd1 + \dps{} ($\lambda=0.1$). We jointly vary the record stride $s \in \{1,2,4,8,16,32\}$ and the last-step delta strategy. Best per-mode \textbf{bolded}. \colorbox{ourscolor}{Shaded} columns indicate our chosen mode.}
\label{tab:stride}
\small
\setlength{\tabcolsep}{4pt}
\newcolumntype{g}{>{\columncolor{ourscolor}}c}
\begin{tabular}{@{}lccccccccgg@{}}
& \multicolumn{2}{c}{\textbf{Raw}} & \multicolumn{2}{c}{\textbf{Neutral}} & \multicolumn{2}{c}{\textbf{Mean}} & \multicolumn{2}{c}{\textbf{Measured}} & \multicolumn{2}{c}{\cellcolor{ourscolor}\textbf{Extrapolate}} \\
\cmidrule(lr){2-3} \cmidrule(lr){4-5} \cmidrule(lr){6-7} \cmidrule(lr){8-9} \cmidrule(lr){10-11}
\textbf{Stride} & \textbf{256} & \textbf{512} & \textbf{256} & \textbf{512} & \textbf{256} & \textbf{512} & \textbf{256} & \textbf{512} & \textbf{256} & \textbf{512} \\
\midrule
1  & \textbf{39.2} & 35.6 & 38.5 & 36.6 & 36.0 & \textbf{40.0} & 37.3 & 39.2 & 38.8 & 40.3 \\
2  & \textbf{39.2} & 36.0 & 38.3 & 35.4 & 37.7 & 39.0 & 38.1 & 39.6 & 37.3 & 39.6 \\
4  & 37.5 & 36.2 & 37.7 & 36.6 & \textbf{37.9} & 39.4 & 36.7 & 40.0 & 38.8 & 40.5 \\
8  & 36.7 & 33.9 & 36.9 & 36.2 & 36.0 & 38.6 & \textbf{37.9} & \textbf{40.2} & \textbf{39.8} & 40.5 \\
16 & 38.8 & \textbf{36.6} & \textbf{39.7} & \textbf{37.1} & 37.8 & 39.0 & 36.7 & 39.6 & 38.5 & 39.6 \\
32 & 38.1 & 35.2 & 37.9 & 36.0 & 37.1 & 39.4 & 37.7 & 39.6 & 38.6 & \textbf{40.9} \\
\midrule
wd1 baseline & \multicolumn{10}{c}{36.7 / 39.6} \\
\midrule
Avg.           & 38.3 & 35.6 & 38.2 & 36.3 & 37.1 & 39.2 & 37.4 & 39.7 & \textbf{38.6} & \textbf{40.2} \\
$\geq$ baseline   & 6/6  & 0/6  & 6/6  & 0/6  & 4/6 & 1/6 & 6/6  & 6/6  & 6/6  & 6/6  \\
\end{tabular}
\end{table}

Extrapolate is the \emph{only} mode that achieves the highest average accuracy at both sequence lengths (38.6 at 256, 40.2 at 512) while exceeding the baseline in every stride configuration (6/6 at both lengths). Raw and Neutral both fail at length 512 (0/6 each), suggesting that scale corruption and information suppression worsen with longer sequences. Measured performs well (6/6 at both lengths) but sacrifices the zero-cost property of \dps{} by requiring an extra forward pass. We therefore adopt Extrapolate as the default last-step mode across all experiments.

Under Extrapolate, every stride $s \in \{1, 2, 4, 8, 16, 32\}$ exceeds the wd1 baseline at both lengths, confirming that the \dps{} signal is reliable across a wide range of recording granularities. The optimal stride varies slightly, but the consistent improvement over the baseline demonstrates that the attribution signal is not an artifact of a particular stride choice.

\subsection{\dps{} normalization mode}
\label{app:ablation-normalization}
 
The \dps{} delta $\delta_k$ (\cref{eq:delta}) produces raw values whose scale varies systematically across the denoising trajectory: early steps operate with little context and yield small deltas, while late steps have rich context and produce larger deltas. Before these deltas enter the weight formula $\omega_i = 1 + \lambda \cdot \bar{\delta}_{\text{birth}(i)}$ (\cref{eq:weight}), they must be normalized to ensure fair credit assignment. The choice of normalization axis---i.e., which cells in the (sample $\times$ step) matrix share a $(\mu, \sigma)$ pair for z-scoring---affects what ``above-average progress'' means. Writing $\delta_{g,k}$ for the delta of sample $g$ at step $k$, we evaluate four modes:
 
\textbf{Per-step} computes a separate mean $\mu_k$ and standard deviation $\sigma_k$ at each denoising step $k$, aggregating across all samples in the batch:
\begin{equation}
    \bar{\delta}_{g,k} = \frac{\delta_{g,k} - \mu_k}{\sigma_k + \epsilon}.
\end{equation}
This automatically handles the scale difference between early and late steps: a delta is ``impressive'' only relative to other samples \emph{at that same step}. The normalization answers: ``At step $k$, which sample made the most progress compared to other samples at this step?''

\textbf{Trajectory} computes a separate $(\mu_g, \sigma_g)$ for each sample $g$, aggregating across all denoising steps:
\begin{equation}
    \bar{\delta}_{g,k} = \frac{\delta_{g,k} - \mu_g}{\sigma_g + \epsilon}.
\end{equation}
This performs no cross-sample comparison---it only identifies which steps were most impactful \emph{within each individual trajectory}. A step receives high weight if it stands out relative to other steps of the same sample, regardless of whether other samples also made large progress at that point. The normalization answers: ``Within this sample's trajectory, which step was the most impactful?''
 
\textbf{Group} computes a single global $(\mu, \sigma)$ across all samples and all steps:
\begin{equation}
    \bar{\delta}_{g,k} = \frac{\delta_{g,k} - \mu_{\text{all}}}{\sigma_{\text{all}} + \epsilon}.
\end{equation}
This is the simplest mode: a delta is above average only if it exceeds the overall mean across the entire batch and all time steps. However, because early and late steps have systematically different magnitudes, late-step deltas tend to dominate the global statistics, potentially under-weighting important early-step transitions. The normalization answers: ``Was this step's progress above or below the overall average across all samples and all times?''
 
\textbf{None} skips normalization entirely, passing raw deltas directly into the weight formula: $\bar{\delta}_{g,k} = \delta_{g,k}$. This makes the modulation sensitive to the absolute log-probability scale, which varies with model architecture, vocabulary size, and sequence length---requiring careful $\lambda$ tuning for each setting.
 
\cref{tab:normalization} ablates the normalization mode on MATH-500 with wd1 + \dps{} ($\lambda = 0.1$), jointly varying the record stride $s \in \{1, 2, 4, 8, 16, 32\}$. We evaluate at generation lengths 256 and 512.
 
\begin{table}[t]
\centering
\caption{\textbf{\dps{} normalization mode ablation} on MATH-500 with wd1 + \dps{} ($\lambda=0.1$). We jointly vary the record stride $s \in \{1,2,4,8,16,32\}$ and the normalization mode. All experiments use the Extrapolate last-step strategy. \colorbox{ourscolor}{Shaded} columns indicate our chosen mode.}
\label{tab:normalization}
\small
\setlength{\tabcolsep}{4pt}
\newcolumntype{g}{>{\columncolor{ourscolor}}c}
\begin{tabular}{@{}lccccccgg@{}}
& \multicolumn{2}{c}{\textbf{Trajectory}} & \multicolumn{2}{c}{\textbf{Group}} & \multicolumn{2}{c}{\textbf{None}} & \multicolumn{2}{c}{\cellcolor{ourscolor}\textbf{Per-step}} \\
\cmidrule(lr){2-3} \cmidrule(lr){4-5} \cmidrule(lr){6-7} \cmidrule(lr){8-9}
\textbf{Stride} & \textbf{256} & \textbf{512} & \textbf{256} & \textbf{512} & \textbf{256} & \textbf{512} & \textbf{256} & \textbf{512} \\
\midrule
1  & 36.4 & 38.8 & 36.0 & 38.3 & 36.4 & 39.0 & 38.8 & 40.3 \\
2  & 36.7 & 38.5 & 37.9 & 39.8 & 37.1 & 38.8 & 37.3 & 39.6 \\
4  & 37.5 & 39.6 & 37.5 & \textbf{41.0} & \textbf{39.0} & 39.4 & 38.8 & 40.5 \\
8  & \textbf{37.9} & 39.4 & 38.3 & 40.3 & 37.5 & 39.4 & \textbf{39.8} & 40.5 \\
16 & 36.7 & 39.4 & 36.4 & 40.9 & 37.9 & 39.8 & 38.5 & 39.6 \\
32 & 36.2 & \textbf{40.2} & \textbf{39.4} & 39.2 & 36.9 & \textbf{40.0} & 38.6 & \textbf{40.9} \\
\midrule
wd1 baseline & \multicolumn{8}{c}{36.7 / 39.6} \\
\midrule
Avg.             & 36.9 & 39.3 & 37.6 & 39.9 & 37.5 & 39.4 & \textbf{38.6} & \textbf{40.2} \\
$\geq$ baseline  & 3/6  & 2/6  & 4/6 & 4/6 & 4/6 & 3/6 & 6/6  & 6/6  \\
\end{tabular}
\end{table}

Per-step is the only mode that exceeds the baseline in all 12 stride--length configurations (6/6 at both lengths) and achieves the highest average accuracy (38.6 / 40.2). The other three modes each fail for a structural reason. Trajectory normalizes within each sample independently, forcing every trajectory to have the same dynamic range regardless of whether it made meaningful progress; this washes out the cross-sample signal \dps{} relies on, and its average at length 512 (39.3) actually falls below the baseline. Group uses a single global $(\mu, \sigma)$, which is dominated by large late-step deltas and suppresses the small but decisive early-step transitions where the model commits to key structural choices. None leaves $\omega_i$ sensitive to the absolute log-probability scale, which drifts with vocabulary size, sequence length, and training stage---making a fixed $\lambda$ unstable across configurations. Per-step avoids all three failures: by computing $(\mu_k, \sigma_k)$ at each step independently, it adapts to the non-stationary delta scale while preserving cross-sample comparisons, and yields modulation factors bounded predictably regardless of training stage. \textbf{We therefore adopt per-step as the default.}

\subsection{\dps{} modulation strength}
\label{app:ablation-lambda}

The modulation strength $\lambda$ in \cref{eq:weight} controls how aggressively \dps{} reweights per-token advantages: $\lambda = 0$ recovers the base method (uniform weighting), while larger values amplify the distinction between decisive and routine denoising steps. Too small a value underutilizes the \dps{} signal; too large a value risks destabilizing training by assigning extreme weights to outlier deltas. \cref{tab:lambda} ablates $\lambda \in \{0.05, 0.1, 0.2\}$ on MATH-500 with wd1 + \dps{}, reporting the best accuracy across all record strides $s \in \{1, 2, 4, 8, 16, 32\}$ (using per-step normalization and Extrapolate last-step mode, as established in \cref{sec:ablation-dps-stride,app:ablation-normalization}).

\begin{table}[t]
\centering
\caption{\textbf{\dps{} modulation strength ablation} on MATH-500 with wd1 + \dps{}. Best accuracy (\%) across record strides $s \in \{1, 2, 4, 8, 16, 32\}$. \colorbox{ourscolor}{Shaded} row indicates our default.}
\label{tab:lambda}
\small
\begin{tabular}{@{}lccc@{}}
$\lambda$ & \textbf{128} & \textbf{256} & \textbf{512} \\
\midrule
0 (baseline) & 32.6 & 36.7 & 39.6 \\
0.05 & 33.4 & 39.2 & \textbf{40.9} \\
\rowcolor{ourscolor} 0.1 & 33.5 & \textbf{39.8} & \textbf{40.9} \\
0.2 & \textbf{35.0} & 39.0 & 40.2 \\
\end{tabular}
\end{table}

All three $\lambda$ values improve over the baseline at every generation length, confirming that \dps{} is beneficial across a range of modulation strengths. An interesting pattern emerges across lengths: $\lambda{=}0.2$ is strongest at $L{=}128$ (+2.4pp over baseline) where aggressive reweighting helps the model focus on the few decisive steps in short outputs, while $\lambda{=}0.1$ is best at $L{=}256$ and tied-best at $L{=}512$ where moderate modulation avoids over-weighting noisy deltas in longer trajectories. We select $\lambda{=}0.1$ as the default since it achieves the best or tied-best accuracy at the two longer generation lengths, which are the more common evaluation settings, while remaining competitive at $L{=}128$.

\subsection{\sml{} sensitivity: number of strata and stratification strategy}
\label{app:ablation-sml}

\begin{figure}[ht]
\centering
\includegraphics[width=\textwidth]{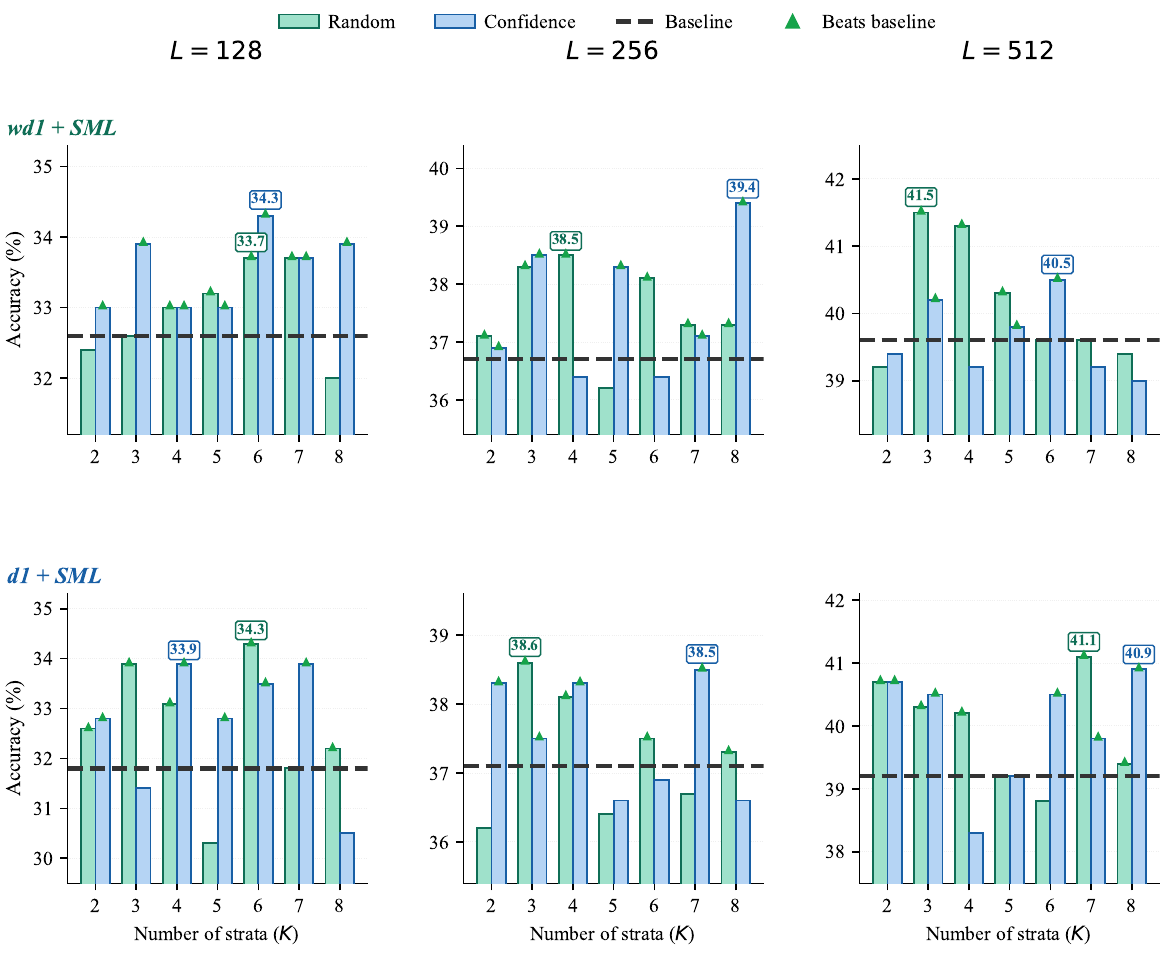}
\caption{\textbf{\sml{} sensitivity on MATH-500 for Diffu-GRPO and wd1.} Grouped bars show accuracy (\%) with random (teal) and confidence-based (blue) stratification for $K \in \{2, \ldots, 8\}$ at three generation lengths. Dashed lines: respective baselines without \sml{}. Green triangles (\textcolor[HTML]{16A34A}{$\blacktriangle$}) mark configurations that meet or exceed the baseline. \sml{} consistently improves wd1 and Diffu-GRPO across most $K$ values, with random stratification showing a smoother profile. 
}
\label{fig:sml-ablation}
\end{figure}

\cref{fig:sml-ablation} ablates the number of strata $K \in \{2, \ldots, 8\}$ and the stratification strategy (random vs.\ confidence-based) on MATH-500 across all d1 and wd1.

\textbf{\sml{} improves wd1 and Diffu-GRPO robustly.} On both methods, the average accuracy across all seven $K$ values exceeds the respective baseline at nearly every generation length. A practitioner choosing $K$ blindly can still expect to see improvements, making \sml{} a low-risk addition. That the pattern holds for both a ratio-free method (wd1) and a ratio-based method (Diffu-GRPO) confirms that the benefit originates from better likelihood estimation rather than from interaction with a specific loss structure.

 \textbf{Random stratification dominates confidence-based.} Confidence-based stratification occasionally produces the single highest accuracy at a given $K$ (e.g., $K{=}6$ on wd1 at $L{=}128$; $K{=}7$ on d1 at $L{=}128$), but its variance across $K$ is substantially higher. This erratic behavior has a principled explanation: confidence-based sorting relies on prediction entropy from a fully-masked forward pass---the same biased regime that \sml{} exists to correct. When entropy estimates are noisy, the resulting strata can be adversarial: hard-to-predict tokens cluster together within a stratum, depriving each other of the contextual benefit that \sml{} is supposed to provide. Random stratification avoids this by ensuring each stratum is a representative sample of the full difficulty spectrum, yielding a smoother profile that requires less per-task tuning.

\subsection{\sml{} regularization weight}
\label{app:ablation-eta}

For ratio-free methods like wd1, \sml{} enters the training loss as an additive regularizer weighted by $\eta$ (\cref{eq:dnc-wd1}). This coefficient balances two objectives: the advantage-weighted base loss (which handles credit assignment between correct and incorrect samples) and the \sml{} likelihood term (which improves token-prediction quality under richer context). Too small an $\eta$ renders \sml{} ineffective; too large an $\eta$ lets the \sml{} term dominate the base loss, reducing the influence of the reward signal.

\begin{table}[ht]
\centering
\caption{\textbf{\sml{} regularization weight ablation} on MATH-500 with wd1 + \sml{}. Best accuracy (\%) across $K \in \{2, \ldots, 8\}$ and both stratification strategies. Best per-length \textbf{bolded}. \colorbox{ourscolor}{Shaded} row indicates our default.}
\label{tab:eta}
\small
\begin{tabular}{@{}lccc@{}}
$\eta$ & \textbf{128} & \textbf{256} & \textbf{512} \\
\midrule
0 (baseline) & 32.6 & 36.7 & 39.6 \\
0.05 & 34.0 & 39.0 & 41.4 \\
\rowcolor{ourscolor} 0.1 & \textbf{34.3} & \textbf{39.4} & \textbf{41.5} \\
0.2 & \textbf{34.3} & 38.6 & 41.1 \\
\end{tabular}
\end{table}

\cref{tab:eta} ablates $\eta \in \{0.05, 0.1, 0.2\}$ on MATH-500 with wd1 + \sml{}, reporting the best accuracy across all strata counts $K \in \{2, \ldots, 8\}$ and both stratification strategies. All three $\eta$ values improve substantially over the baseline at every generation length (+1.4--1.9pp average gain), confirming that \sml{} is beneficial across a wide range of regularization strengths. The spread between worst and best $\eta$ at any given length is small: 0.3pp at $L{=}128$, 0.8pp at $L{=}256$, and 0.4pp at $L{=}512$. This insensitivity is a practical advantage---a practitioner can adopt $\eta{=}0.1$ without per-task tuning. We select $\eta{=}0.1$ as the default since it achieves the best or tied-best accuracy at all three generation lengths.

\subsection{Computational overhead}
\label{app:compute-overhead}

All experiments are conducted on a cluster of 8$\times$ NVIDIA H100 80GB GPUs using QLoRA (4-bit quantization with LoRA rank 128). We measure the overhead of \dps{} and \sml{} independently on top of the wd1 base method, running 100 training steps with the default configuration (\cref{tab:hyperparams-training}).

\begin{table}[ht]
\centering
\caption{\textbf{Computational overhead of \dps{} and \sml{}.} Wall-clock time measured over 100 training steps on wd1 with LLaDA-8B-Instruct. Overhead (\%) is relative to the wd1 baseline.}
\label{tab:overhead}
\small
\begin{tabular}{@{}lrr@{}}
\textbf{Configuration} & \textbf{Wall-clock (s)} & \textbf{Overhead (\%)} \\
\midrule
\multicolumn{3}{l}{\textit{Baseline}} \\
wd1 & 3\,768 & --- \\
\midrule
\multicolumn{3}{l}{\textit{\dps{} (varying record stride $s$)}} \\
\quad $s = 1$  & 3\,887 & +3.2 \\
\quad $s = 4$  & 3\,797 & +0.8 \\
\quad $s = 16$ & 3\,785 & +0.5 \\
\quad $s = 32$ & 3\,771 & +0.1 \\
\midrule
\multicolumn{3}{l}{\textit{\sml{} (varying number of strata $K$)}} \\
\quad $K = 2$ & 4\,014 & +6.5 \\
\quad $K = 4$ & 4\,208 & +11.7 \\
\quad $K = 8$ & 4\,590 & +21.8 \\
\end{tabular}
\end{table}

\cref{tab:overhead} presents the wall-clock time across varying \dps{} record strides and \sml{} strata counts and reveals two patterns.

\textbf{\dps{} is near-zero cost.} Wall-clock overhead scales linearly with the number of recorded snapshots ($T_{\text{diff}} / s$): from +0.1\% at $s{=}32$ (4 snapshots per trajectory) to +3.2\% at $s{=}1$ (128 snapshots). At our recommended operating range of $s \in \{4, 8, 16, 32\}$, the overhead is under 1\%---negligible relative to the $G \times T_{\text{diff}} = 6 \times 128 = 768$ generation-phase forward passes that dominate each training step.

\textbf{\sml{} scales linearly with $K$.} Wall-clock overhead is +6.5\% at $K{=}2$, +11.7\% at $K{=}4$ (our default), and +21.8\% at $K{=}8$, consistent with $K$ additional forward passes per inner iteration during loss computation. Even at $K{=}4$, the overhead is modest: loss computation accounts for a small fraction of each training step compared to the generation phase. Since \dps{} and \sml{} operate in different phases of the training loop---\dps{} during generation, \sml{} during loss computation---their overheads are additive and non-interacting.

\subsection{Constrained decoding: JSON generation}
\label{app:json-generation}

\begin{table}[ht]
\centering
\caption{\textbf{JSON generation results.} Schema Adherence Rate (\%) on \textsc{github-medium}. Diffu-GRPO and wd1 evaluated at $L{=}512$; GDPO at $L{=}256$. Best accuracy across record strides (for \dps{}) or strata counts and stratification strategies (for \sml{}) reported. \textcolor{deltagreen}{Green} values show absolute change over the reproduced baseline. \colorbox{ourscolor}{Blue rows} are our methods.}
\label{tab:json-results}
\small
\begin{tabular}{@{}lcl@{}}
\textbf{Method} & \textbf{Schema} & \\
 & \textbf{Adherence (\%)} & \\
\midrule
LLaDA-8B-Instruct ($L{=}256$) & 48.83 & \\
LLaDA-8B-Instruct ($L{=}512$) & 60.55 & \\
\midrule
Diffu-GRPO ($L{=}512$) & 69.44 & \\
\rowcolor{ourscolor}
\quad + \dps{} & \textbf{73.96} & {\scriptsize\color{deltagreen}\textbf{+4.52}} \\
\rowcolor{ourscolor}
\quad + \sml{} & 71.18 &  {\scriptsize\color{deltagreen}+1.74}\\
\rowcolor{ourscolor}
\quad + \ours{} & 73.61 &  {\scriptsize\color{deltagreen}+4.17}\\
\midrule
wd1 ($L{=}512$) & 72.57 & \\
\rowcolor{ourscolor}
\quad + \dps{} & 76.04 & {\scriptsize\color{deltagreen}+3.47} \\
\rowcolor{ourscolor}
\quad + \sml{} & 76.39 & {\scriptsize\color{deltagreen}+3.82} \\
\rowcolor{ourscolor}
\quad + \ours{} & \textbf{78.47} & {\scriptsize\color{deltagreen}\textbf{+5.90}} \\
\midrule
GDPO ($L{=}256$) & 63.19 & \\
\rowcolor{ourscolor}
\quad + \dps{} & 64.23 & {\scriptsize\color{deltagreen}+1.04} \\
\rowcolor{ourscolor}
\quad + \sml{} & 65.28 & {\scriptsize\color{deltagreen}+2.09} \\
\rowcolor{ourscolor}
\quad + \ours{} & \textbf{67.01} & {\scriptsize\color{deltagreen}\textbf{+3.82}}
\end{tabular}
\end{table}

\cref{tab:json-results} reports Schema Adherence Rate on the \textsc{github-medium} JSON benchmark, which is the fraction of outputs that conform to the target schema. Note that the output conforming to the schema implies that it also parses as valid JSON. We evaluate Diffu-GRPO and wd1 at $L{=}512$ and GDPO at $L{=}256$ (since the implementation of GDPO for the JSON task runs into OOM issues with $L{=}512$). For each method we report the best accuracy across all record strides $s \in \{1, 2, 4, 8, 16, 32\}$ (for \dps{}) or strata counts and stratification strategies (for \sml{}).

\textbf{Analysis.} Two patterns are notable. First, \ours{} achieves the highest improvement on wd1 (+5.90pp) and GDPO (+3.82pp), with both \dps{} and \sml{} contributing complementary gains---on wd1, \dps{} alone adds +3.47pp and \sml{} alone adds +3.82pp, while the combination exceeds both at +5.90pp. On d1, \dps{} alone (+4.52pp) slightly outperforms \ours{} (+4.17pp), suggesting mild interference between the two components under the ratio-based formulation for this task.

Second, unlike the short-output constraint satisfaction tasks (Countdown, Sudoku) where \sml{} caused mode collapse (Section~\ref{sec:sml-limitations}), \sml{} is effective on JSON generation despite JSON being a structurally constrained format. This confirms that \sml{}'s failure on constraint tasks is driven by \emph{output length}, not by the presence of structural constraints: JSON outputs at $L{=}512$ are long enough that revealing 75\% of the sequence does not trivially determine the remaining tokens, so the \sml{} gradient teaches useful predictive skills that transfer to generation. This result sharpens the applicability boundary of \sml{}: it is effective whenever the output is sufficiently long and semantically diverse, regardless of whether the task imposes structural constraints.

\subsection{Training Dynamics: \dps{} Accelerates Reward Growth}
\label{sec:training-dynamics}

\begin{figure}[ht]
\centering
\includegraphics[width=\linewidth]{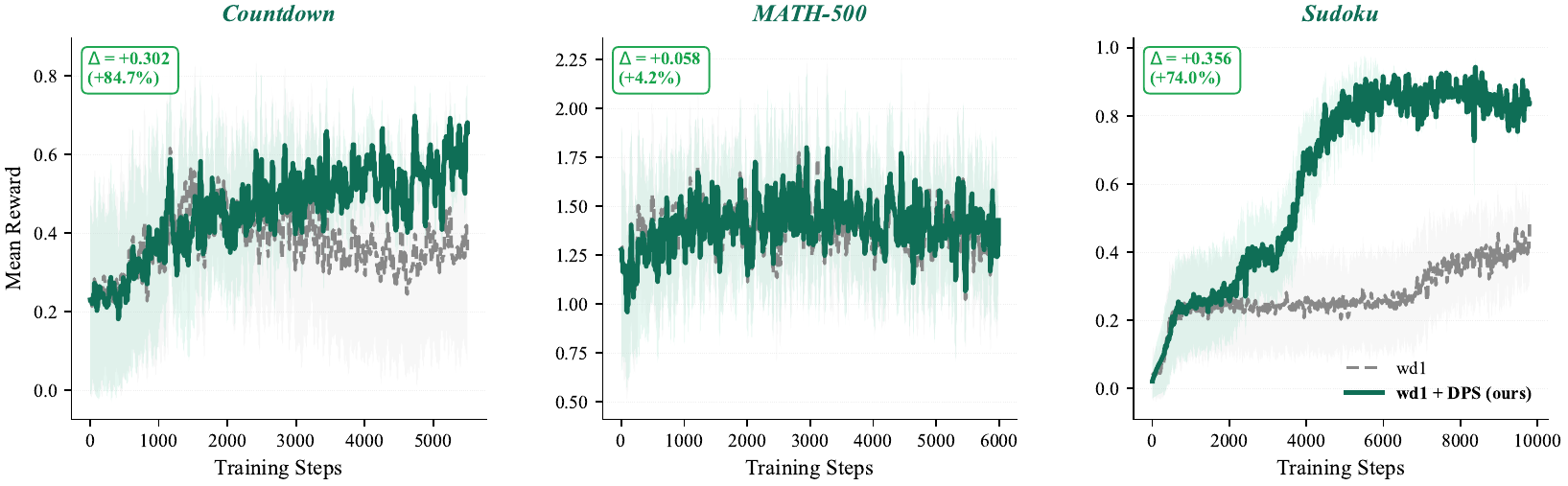}
\caption{\textbf{Training dynamics: \texttt{wd1} vs \texttt{wd1\,+\,\dps{}}.} Mean reward (solid lines) with $\pm 1$ group std (shaded bands), smoothed over a 20-step rolling window. \dps{} produces an immediate and persistent gap on the structured benchmarks (Countdown, Sudoku) and a smaller but consistent gain on MATH-500. The $\Delta$ box reports the absolute and relative final-reward improvement. All runs use identical hyperparameters except for the \dps{} modulation.}
\label{fig:reward-curves}
\end{figure}

To understand \emph{when} during training \dps{} contributes, we compare reward trajectories of \texttt{wd1} and \texttt{wd1\,+\,\dps{}} on three representative benchmarks: Countdown, MATH-500, and Sudoku (\cref{fig:reward-curves}).
Reward is the per-step group mean, smoothed with a 20-step rolling window; shaded bands denote $\pm$ one group standard deviation. Each pair shares the same seed, batch size, generation length, and reward functions, so the only difference is whether the per-token advantage is modulated by $\bar{\delta}_{\mathrm{birth}(i)}$.

\textbf{\dps{}} On Countdown and Sudoku, the \texttt{wd1\,+\,\dps{}} curves separate from the baseline within the first few hundred steps and continue widening throughout training, ending at $+0.302$ ($+84.7\%$) and $+0.356$ ($+74.0\%$) absolute mean reward respectively. Both tasks have tight inter-token dependencies (arithmetic operators, grid constraints) so \emph{which} denoising step commits the model to a particular structural choice matters disproportionately—the precise signal that uniform credit assignment dilutes and that \dps{} recovers.

\textbf{On MATH-500 the gap is smaller but consistent.} The two curves co-evolve early on—both methods learn the same easy reward shaping—but \texttt{wd1\,+\,\dps{}} pulls ahead in the second half of training and ends $+0.058$ ($+4.2\%$) above baseline. Long, free-form math completions contain more routine fill-in tokens (whitespace, repeated digits, formatting), so the fraction of steps with high $\delta_k$ is lower; the residual gain comes exclusively from the small subset of decisive steps that \dps{} up-weights. This is consistent with our results in \cref{tab:main-math-code}, where SML—targeting the orthogonal weakness of biased likelihood estimates—provides the complementary gains on long-form math reasoning.

\textbf{Stability.} Across all three datasets, \texttt{wd1\,+\,DPS} exhibits
no additional variance: the shaded bands track the baseline closely and the
mean curve is monotone after smoothing. This supports our claim that DPS is
a \emph{plug-and-play} modification—it modulates the existing advantage
rather than introducing new gradient sources, so it cannot destabilise the
trust region or amplify reward variance.

\applefootnote{\textcolor{textgray}{\sffamily Apple and the Apple logo are trademarks of Apple Inc., registered in the U.S. and other countries and regions.}}

\end{document}